\documentclass[letterpaper, 10 pt, journal, twoside]{IEEEtran}

\usepackage{amsmath,amsfonts,amssymb}
\usepackage{array}
\usepackage[caption=false,font=normalsize,labelfont=sf,textfont=sf]{subfig}
\usepackage{textcomp}
\usepackage{stfloats}
\usepackage{url}
\usepackage{verbatim}
\usepackage{graphicx}
\usepackage{cite}
\usepackage{comment}
\usepackage[hidelinks]{hyperref}
\hypersetup{
   linkcolor=blue,
   breaklinks=true,   % splits links across lines
   colorlinks=true,   % displays links as colored text instead of blocks
   citecolor=blue,
   urlcolor=blue
}
\usepackage{siunitx,booktabs}
\usepackage{multirow}
\usepackage[table,xcdraw]{xcolor}
\usepackage{colortbl}
\usepackage{hhline}
\usepackage[font=footnotesize,labelfont=bf,tableposition=top]{caption}
\usepackage{soul}
\usepackage{graphicx}

\definecolor{hidden-draw}{RGB}{205, 44, 36}
\definecolor{lightgray}{gray}{0.9}
\definecolor{lightblue}{RGB}{173, 216, 230}

\usepackage[ruled,vlined,linesnumbered]{algorithm2e}

\SetCommentSty{mycommfont}
\SetKwComment{Comment}{$\triangleright$\ }{}
\let\oldnl\nl%
\newcommand{\nonl}{\renewcommand{\nl}{\let\nl\oldnl}}%

\usepackage{titlesec}
\titlespacing*{\subsection}{0pt}{0.1cm}{0.1cm}

\usepackage[most]{tcolorbox}
\usepackage[T1]{fontenc}

\usepackage{listings}
\usepackage{xcolor}
\definecolor{codegreen}{rgb}{0,0.6,0}
\definecolor{codegray}{rgb}{0.5,0.5,0.5}
\definecolor{codepurple}{rgb}{0.58,0,0.82}
\definecolor{backcolour}{rgb}{0.95,0.95,0.92}  % final version

\lstdefinestyle{mystyle}{
    backgroundcolor=\color{blue!5!white},  % Change to the backcolor above in the final version 
    commentstyle=\color{codegreen},
    keywordstyle=\color{magenta},
    numberstyle=\tiny\color{codegray},
    stringstyle=\color{codepurple},
    basicstyle=\ttfamily\scriptsize,
    breakatwhitespace=false,         
    breaklines=true,                 
    captionpos=b,                    
    keepspaces=true,                 
    showspaces=false,                
    showstringspaces=false,
    showtabs=false,                  
    tabsize=2
}

\newcolumntype{C}[1]{>{\centering\arraybackslash}p{#1}}
\newcolumntype{P}[1]{>{\raggedright\arraybackslash\noindent}p{#1}}

\definecolor{redish}{RGB}{255, 211, 211}
\definecolor{greenish}{RGB}{194, 241, 200}

\usepackage{censor}
\def\censorcolor{gray!50} \let\svcensorrule\censorrule \renewcommand\censorrule[1]{ \textcolor{\censorcolor}{\svcensorrule{#1}}}

\usepackage{xcolor}

\usepackage[switch]{lineno}

\newcommand \name{MotionBenchGen\xspace}

\begin{document}

% Paper sections
\title{\huge{\textbf{PerFACT}: Motion \underline{\textbf{P}}olicy with LLM-Pow\underline{\textbf{er}}ed Dataset Synthesis and \underline{\textbf{F}}usion \underline{\textbf{A}}ction-\underline{\textbf{C}}hunking \underline{\textbf{T}}ransformers}}

\author{Davood Soleymanzadeh$^{1}$, Xiao Liang$^{2}$, and Minghui Zheng$^{1}$
        % <-this % stops a space
\thanks{$^{1}$Davood Soleymanzadeh and Minghui Zheng are with the J. Mike Walker '66 Department of Mechanical Engineering, Texas A\&M University, College Station, TX 77843, USA (\tt\footnotesize e-mail: davoodso@tamu.edu; mhzheng@tamu.edu).}% <-this % stops a space
\thanks{$^{2}$Xiao Liang is with the Zachry Department of Civil and Environmental
Engineering, Texas A\&M University, College Station, TX 77843 USA (\tt\footnotesize e-mail:
xliang@tamu.edu).}
\thanks{This work was supported by the USA National Science Foundation under Grant No. 2026533/2422826. Portions of this research were conducted with the advanced computing resources provided by Texas A\&M High Performance Research Computing.}
}
\maketitle
\begin{abstract}
Deep learning methods have significantly enhanced motion planning for robotic manipulators by leveraging prior experiences within planning datasets. However, state-of-the-art neural motion planners are primarily trained on small datasets collected in manually generated workspaces, limiting their deployment in various everyday scenarios. Additionally, these planners often rely on monolithic network architectures that struggle to encode critical planning information. To address these challenges, we introduce Motion Policy with Dataset Synthesis powered by large language models (LLMs) and Fusion Action-Chunking Transformers (PerFACT), which incorporates two key components. Firstly, a novel workspace generation method, \name, enables large-scale planning data collection by leveraging procedural primitive generation, and LLM-powered primitive suggestion and placement. Secondly, we introduce Fusion Motion Policy Networks (M$\pi$NetsFusion), an end-to-end, open-loop neural motion planner that uses a fusion action-chunking transformer to better encode planning signals and attend to multiple feature modalities. Leveraging \name, we collect a dataset of 3.5M trajectories to train and evaluate M$\pi$NetsFusion against state-of-the-art planners. Results show that M$\pi$NetsFusion achieves consistently low planning time with sub-second inference, while maintaining competitive performance compared to both sampling-based and end-to-end neural benchmark planners. Project website: \href{https://davoodsz.github.io/perfact.github.io/}{https://davoodsz.github.io/perfact.github.io/}
\end{abstract}

\section{Introduction} \label{sec: intro}
\IEEEPARstart{M}{otion} planning problem is a crucial component of the robotic manipulation stack, which involves finding a collision-free path between a start and goal configuration \cite{lavalle2006planning}. Traditional methods, like sampling-based \cite{orthey2023sampling} and trajectory optimization \cite{zucker2013chomp} techniques, plan motions from scratch for each new problem, which consume a significant amount of computational resources and require privileged workspace knowledge.

Neural motion planners have been developed to enhance the performance of traditional motion planning algorithms \cite{soleymanzadeh2025simpnet}. However, these planners are often trained on limited datasets collected in manually designed workspaces, which limits their ability to generalize to out-of-distribution planning scenarios. This limitation is especially significant in motion planning for robotic manipulators, given the inherent discontinuities within the motion planning problems \cite{farber2003topological,chamzas2021learning}. As a result, large-scale datasets are required to establish general-purpose neural motion planners. 

To address this limitation, procedural workspace generation \cite{chamzas2021motionbenchmaker, dalal2024neural} employ basic shapes, and primitives to generate everyday objects of various sizes and configurations. Then, collision-aware random placement is leveraged to generate diverse workspaces for planning data collection. However, using random placement limits the diversity and semantic feasibility of the generated workspaces. More broadly, LLMs have been leveraged for scene and task generation within the manipulation autonomy stack \cite{wang2024robogen, zhao2023survey, wang2024survey}. These models are also being used to synthesize scenarios for robotic manipulation and locomotion \cite{wang2024robogen}. In these scenarios, articulated objects are either sampled from existing 3D datasets \cite{deitke2023objaverse}, or are generated by state-of-the-art generative AI tools \cite{wang2023score}. However, existing 3D datasets lack the diversity of everyday articulated objects in terms of shape and configuration, and generative AI tools also struggle to generate everyday objects with a high level of diversity.

\begin{figure*}[t] 
\centering
\includegraphics[width=0.9\textwidth]{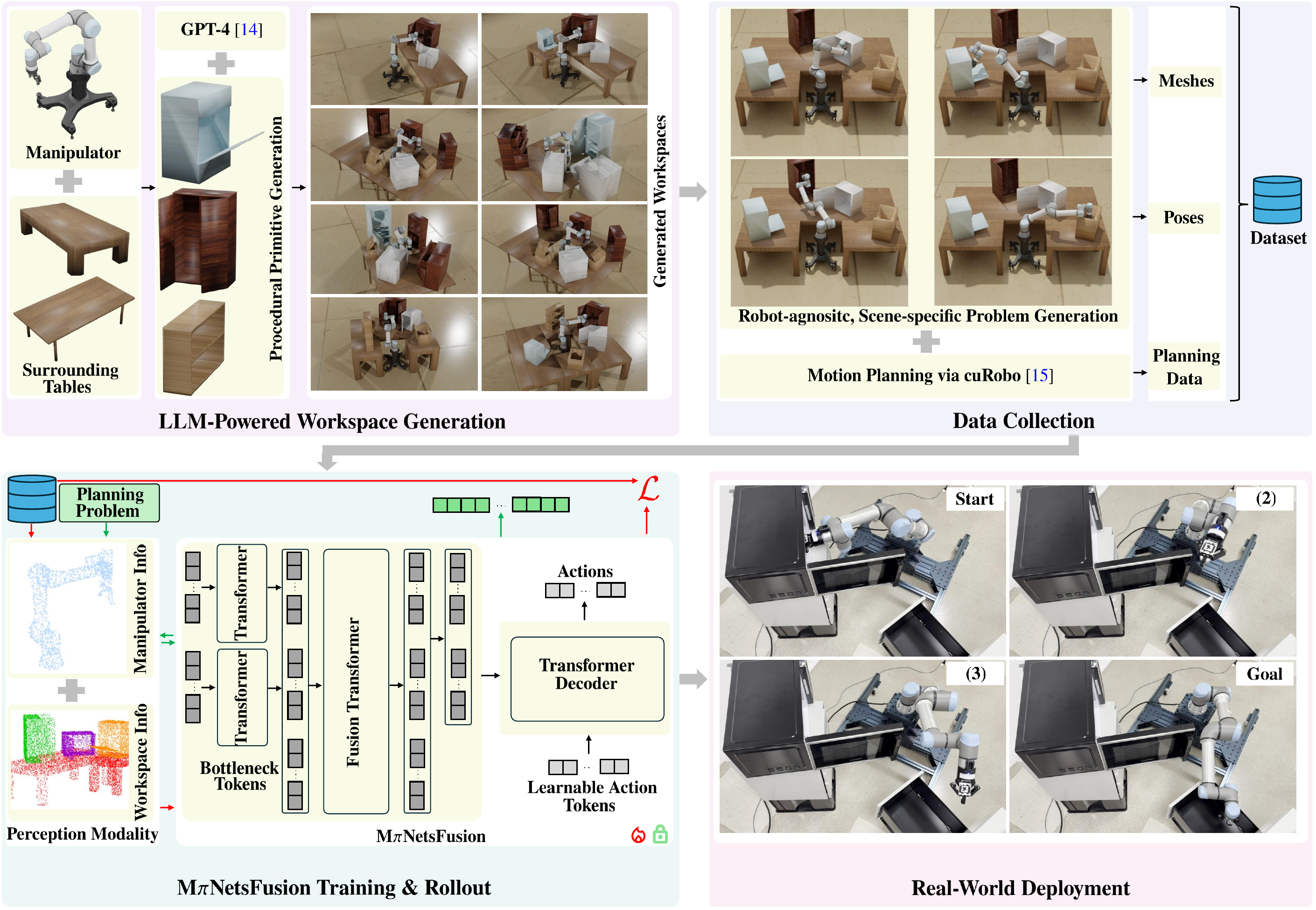}
\caption{\textbf{PerFACT's} various components and real-world deployment. Firstly, \name (Section \ref{sec: motiongeneralizer}) integrates procedural primitive generation with a large language model's (LLM) -\textbf{GPT-4} \cite{achiam2023gpt}- reasoning capabilities to produce diverse workspaces for training generalist neural motion planners. Then, \name generates robot-agnostic, scene-specific motion-planning problems (Section \ref{subsec: planningproblems}) and integrates with state-of-the-art motion planners (\textbf{cuRobo} \cite{sundaralingam2023curobo}) to generate a large-scale planning dataset. The dataset, combined with the perception modality (Section \ref{subsec: modality}) from \name, is leveraged to \colorbox{white}{train the M$\pi$NetsFusion} (Section \ref{sec: motionplanner}). The sensing modality (Section \ref{subsec: modality}) is also utilized within the \colorbox{white}{open-loop rollout of M$\pi$NetsFusion} to solve motion-planning problems in held-out evaluation scenarios (Section \ref{sec:main_eval}). Finally, M$\pi$NetsFusion is deployed within real-world planning scenarios (Section \ref{sec: experiments}). The path profile from start configuration to the goal configuration is demonstrated in frames 2-3.}
\label{fig_1- motiongeneralizer_overview}
\end{figure*}

Additionally, state-of-the-art neural motion planners typically utilize current and goal robot configuration embeddings and workspace embeddings to predict the next planning state toward the goal, either for informed sampling \cite{qureshi2019motion} or end-to-end open-loop planning \cite{fishman2023motion, dalal2024neural}. These frameworks typically utilize simple feature concatenation to fuse and process configuration space (i.e., manipulator information) and workspace features (i.e., obstacle information), and perform feature-level fusion for motion planning \cite{qureshi2019motion, fishman2023motion, dalal2024neural}. This makes the learning process biased towards dominant planning modalities (e.g., workspace embeddings) and discards essential, yet sparse, planning signals (e.g., current and goal configuration embeddings).

In this work, we propose PerFACT, which contains two critical components to address the limitations of neural motion planners for robotic manipulators:

\begin{itemize}
    \item \textbf{\name} (see Figure \ref{fig_1- motiongeneralizer_overview}) provides diverse, feasible workspaces for large-scale planning data collection.
    \item \textbf{M$\pi$NetsFusion} (See Figure \ref{fig_1- motiongeneralizer_overview}) utilizes motion planning modality-aware network architectures to attend to different planning feature modalities.
\end{itemize}

\name integrates procedural primitive generation with the reasoning capabilities of LLMs to generate diverse planning workspaces. Procedural primitive generation is used to randomly generate various everyday articulated objects with different shapes and configurations, while an LLM is queried in a chain-of-thought (CoT) manner to get the number, position, and orientation of procedurally generated primitives for planning workspace generation.

The following are the contributions of \name:

\begin{enumerate}
    \item \name adopts procedural primitive generation frameworks from Neural MP \cite{dalal2024neural}, and NVIDIA SceneSynthesizer \cite{eppner2025scene_synthesizer} to generate everyday object primitives. Then, it prompts an LLM in a CoT manner to get the number of primitives per scene, and their corresponding poses to generate planning workspaces.
    
    \item \name generates a synthetic workspace point cloud by uniformly sampling points on obstacle meshes within the workspace, and on the robotic manipulator meshes at different configurations. It also generates robot-agnostic, scene-specific motion planning problems by sampling collision-free poses within workspace primitives and the robotic manipulator's reachable region. State-of-the-art motion planners, such as OMPL \cite{sucan2012open} and cuRobo \cite{sundaralingam2023curobo}, can be used to collect planning data given these problems.
\end{enumerate}

M$\pi$NetsFusion leverages a fusion transformer architecture to attend to different planning modalities and learn their inter-communications. Specifically, it utilizes bottleneck transformers as the encoder within Action Chunking Transformers (ACT) \cite{zhao2023learning} framework to encode these cross-modal interactions.

The following are the contributions of M$\pi$NetsFusion:
\begin{enumerate}
    \item M$\pi$NetsFusion employs bottleneck transformers \cite{nagrani2021attention, higuera2025tactile} to model the contribution of each planning modality. The fusion encoder restricts information flow to distill dense workspace embeddings and sparse configuration data into a compact representation.
\end{enumerate}

The paper is organized as follows. Section \ref{sec: related-work} reviews related work on scene generation for planning data collection and neural motion planning for robotic manipulators. Section \ref{sec: motiongeneralizer} presents \name, describes its components, and evaluations. Section \ref{sec: motionplanner} introduces M$\pi$NetsFusion. Section \ref{sec:main_eval} evaluates and compares M$\pi$NetsFusion with state-of-the-art motion planners. Section \ref{sec: experiments} demonstrates real-world deployment of the proposed neural motion planner. Section \ref{sec: conclusion} concludes the paper.

\section{Related Work} \label{sec: related-work}
\noindent
\textbf{Workspace Generation for Motion Planning.} Roboticists have developed several frameworks to generate diverse and large-scale workspaces/environments for learning motion and manipulation policies~\cite{liu2023meshdiffusion, long2024wonder3d, hu2025simulating}. However, generated objects with these methods lack the modular composability and articulation required for training neural motion planners. Also, existing 3D object datasets - e.g., Objaverse \cite{deitke2023objaverse} and PartNet-Mobility \cite{xiang2020sapien} - have been utilized for robotic manipulation \cite{nasiriany2024robocasa}. However, these datasets lack articulated objects with diverse sizes and articulations required for motion planning.

\begin{figure*}[htbp] 
\centering
\includegraphics[width=\textwidth]{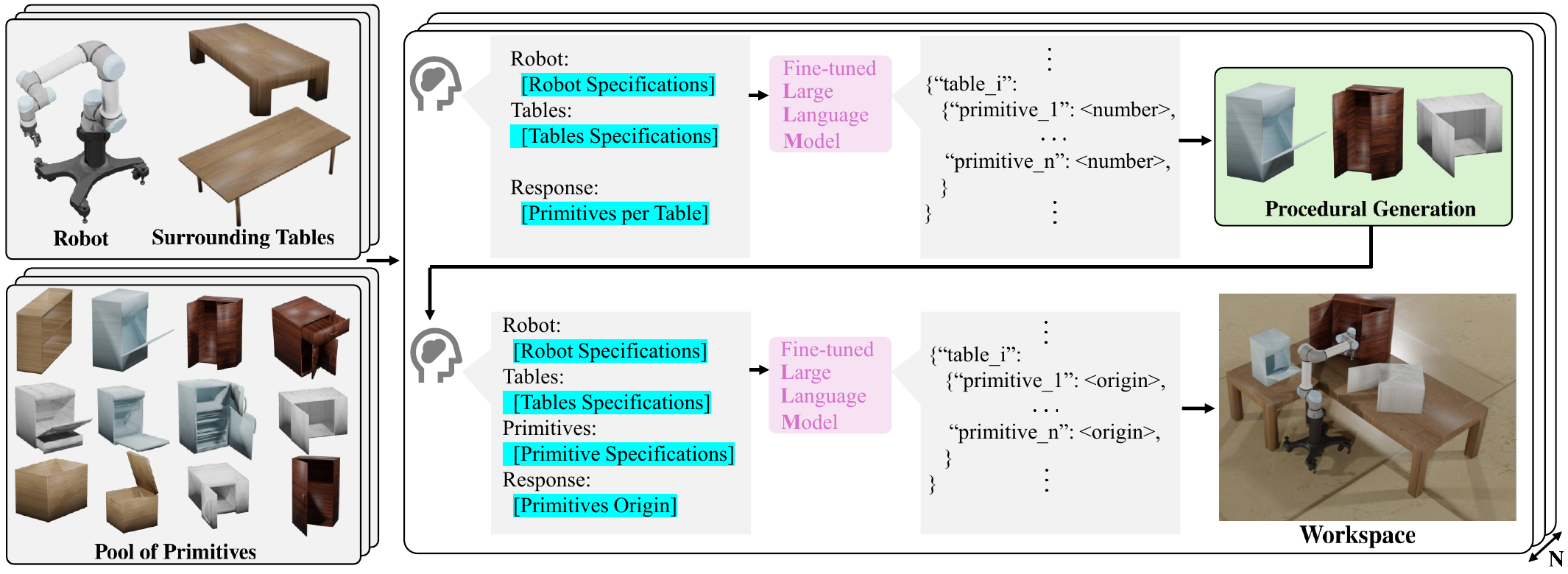}
\caption{\textbf{\name} framework for diverse workspace generation. This method first randomly selects a robot type and the number of its surrounding tables, which are created using procedural generation. Next, it prompts the LLM in a few-shot manner (\textbf{Fine-tuned Large Language Model}) to determine the number of primitives for each table. The suggested primitives are then procedurally generated, given the pool of primitives, and the LLM is prompted again in a few-shot manner (\textbf{Fine-tuned Large Language Model}) to specify the location and orientation of each primitive on its respective table, and outputs the workspace. The procedure can be repeated to generate a diverse set of planning workspaces with an arbitrary number (\textbf{N}).}
\label{fig_2- motiongeneralizer-framework}
\end{figure*}

Procedural primitive generation methods create everyday objects programmatically and place them randomly within the manipulator's workspace for motion planning~\cite{chamzas2021motionbenchmaker, fishman2023motion}. However, the resulting workspaces are not cluttered, as they only contain one major obstacle, and the workspace variation is limited. Neural MP \cite{dalal2024neural} addresses the limitations of procedural primitive generation methods~\cite{chamzas2021motionbenchmaker, fishman2023motion} by generating everyday objects with diverse sizes and shapes. It creates these objects by combining cuboid primitives and randomly placing them in the scene for workspace generation. However, the generated workspaces may not seem semantically realistic due to the random placement procedure. \name combines procedural primitive generation with high-level reasoning of large language models (LLMs) to create large-scale datasets for neural motion planners for robotic manipulators. 

\vspace{0.2cm}
\noindent
\textbf{Language Models in Robotics.} LLMs' semantic reasoning, problem-solving, and visual interpretation capabilities~\cite{achiam2023gpt, kirillov2023segment} have proven useful for establishing generalist robots \cite{li2022grounded, chen2024autotamp}. However, these models are mainly used for high-level robotic planning because they mainly reason about semantics and textual prompts. LLMs' application within low-level planning and control relies on additional auxiliary components, such as motion primitives \cite{liang2023code} and motion optimizers \cite{huang2023voxposer}, since these models are not inherently trained on physical interactions and trajectories \cite{kwon2024language}.

\vspace{0.2cm}
\noindent
\textbf{Neural Motion Planning.} Deep learning frameworks have been widely used to improve motion planning for robotic manipulators. These methods either enhance specific components of classical planners \cite{soleymanzadeh2025simpnet, qureshi2019motion} or function as end-to-end, open-loop planners \cite{fishman2023motion, dalal2024neural, yang2025deep}. They have been employed as informed samplers \cite{lai2021plannerflows, johnson2023learning, johnson2024zero, soleymanzadeh2025simpnet} and collision checkers \cite{song2024graph, kim2022graphdistnet, krawczyk2023comparison, murali2023cabinet} within sampling-based algorithms, used to learn constrained manifolds for constrained motion planning \cite{park2024constrained, lembono2021learning}, and applied to warm-start trajectory-optimization methods \cite{ichnowski2020deep}. Recently, the trajectory optimization formulation has been reformulated as a planning-as-inference problem, where diffusion models~\cite{chi2025diffusion, yang2024equibot} are used to generate trajectories through guided denoising that enforces planning constraints~\cite{carvalho2023motion, yan2025m, saha2024edmp}. However, most existing approaches are trained on relatively small datasets derived from tabletop environments. \name provides a workspace-generation workflow that enables the creation of large-scale, diverse datasets for training neural motion planners.

\begin{algorithm}[h]
\caption{\name}
\label{alg:motiongeneralizer}
\DontPrintSemicolon
\KwIn{
Number of workspaces $N$; manipulator $\mathrm{Manip}$; number of tables $N_{\mathrm{tab}}$; LLM~\cite{achiam2023gpt}; primitive generators from NeuralMP~\cite{dalal2024neural} and SceneSynthesizer~\cite{eppner2025scene_synthesizer}; collision checker $Q$; inverse kinematics $\mathrm{IK}$; point-cloud sampler $\mathrm{PCDSampler}$.
}

\KwOut{
Generated workspaces $\mathcal{W}$, workspace point clouds $\mathcal{P}$, and planning problems $\mathcal{G}$.
}

$\mathcal{W} \leftarrow \emptyset$, $\mathcal{P} \leftarrow \emptyset$, $\mathcal{G} \leftarrow \emptyset$\;

\For{$i \leftarrow 1$ \KwTo $N$}{
    $\mathrm{Workspace} \leftarrow \emptyset$\;

    \tcp{Generate scene structure}
    $\mathrm{Tables} \leftarrow \mathrm{GenerateTables}(N_{\mathrm{tab}}, \mathrm{Manip})$\;
    $\mathrm{TablePrimitives} \leftarrow \mathrm{LLM}(\mathrm{Manip}, \mathrm{Tables})$\;
    $\mathrm{Primitives} \leftarrow \mathrm{GeneratePrimitives}(\mathrm{TablePrimitives})$\;
    $\mathrm{PrimitivePoses} \leftarrow \mathrm{LLM}(\mathrm{Tables}, \mathrm{Primitives}, \mathrm{Manip})$\;

    \tcp{Validate and construct workspace}
    \For{each $\mathrm{Primitive}$ in $\mathrm{Primitives}$}{
        \tcp{Assign pose from $\mathrm{PrimitivePoses}$ to $\mathrm{Primitive}$}

        \While{$Q(\mathrm{Workspace}, \mathrm{Primitive})$}{
            \tcp{Update primitive pose using collision-normal correction}
        }

        $\mathrm{Workspace} \leftarrow \mathrm{Workspace} \cup \{\mathrm{Primitive}\}$\;
    }

    $\mathcal{W} \leftarrow \mathcal{W} \cup \{\mathrm{Workspace}\}$\;

    \tcp{Generate perception modality}
    $\mathrm{WorkspacePCD} \leftarrow \emptyset$\;
    \For{each $\mathrm{Primitive}$ in $\mathrm{Workspace}$}{
        $\mathrm{PrimitivePCD} \leftarrow \mathrm{PCDSampler}(\mathrm{Primitive})$\;
        $\mathrm{WorkspacePCD} \leftarrow \mathrm{WorkspacePCD} \cup \{\mathrm{PrimitivePCD}\}$\;
    }
    $\mathcal{P} \leftarrow \mathcal{P} \cup \{\mathrm{WorkspacePCD}\}$\;

    \tcp{Generate planning problems}
    $\mathcal{G}_{\mathrm{ws}} \leftarrow \emptyset$\;
    \For{each $\mathrm{Primitive}$ in $\mathrm{Workspace}$}{
        $\mathrm{Pose} \leftarrow \mathrm{PoseSampler}(\mathrm{Primitive})$\;
        $\mathrm{Config} \leftarrow \mathrm{IK}(\mathrm{Pose})$\;

        \If{$Q(\mathrm{Workspace}, \mathrm{Config}) = \mathrm{False}$}{
            $\mathcal{G}_{\mathrm{ws}} \leftarrow \mathcal{G}_{\mathrm{ws}} \cup \{\mathrm{Config}\}$\;
        }
    }
    $\mathcal{G} \leftarrow \mathcal{G} \cup \{\mathcal{G}_{\mathrm{ws}}\}$\;
}

\Return $\mathcal{W}, \mathcal{P}, \mathcal{G}$\;
\end{algorithm}

\vspace{0.2cm}
\noindent
\textbf{Neural Motion Planners' Architectures:} The success of neural motion planners began with Motion Planning Networks (MPNets) and its variants \cite{qureshi2019motion, qureshi2020motion, qureshi2020neural, qureshi2021constrained}, where the robot's current configuration, goal configuration, and workspace embeddings are concatenated as input features to a neural network that predicts the next time step configuration. Subsequent works, such as Motion Policy Networks (M$\pi$Nets), Neural MP \cite{dalal2024neural}, and Deep Reactive Policy \cite{yang2025deep} have extended this paradigm by leveraging a similar formulation to predict the next time step configuration within an end-to-end open loop planning framework. These frameworks typically fuse multiple planning sensing modalities via direct concatenation to predict the next time-step configurations towards the planning goal. However, since these global motion planners are mainly developed for static environments, the workspace point cloud often becomes the dominant and constant feature across diverse planning instances, which affects the performance of neural motion planners.

To address this limitation, Spatial-informed Motion Planning Networks (SIMPNet) \cite{soleymanzadeh2025simpnet} utilized a cross attention mechanism to attend to the workspace embedding, allowing the network to focus on other sparse yet critical planning modalities. Also, other architectures have explored attention mechanisms to improve efficiency and information flow in high-dimensional multimodal inputs. Perceiver~\cite{jaegle2021perceiver} and Perceiver IO~\cite{jaegle2021perceiverio} employ latent arrays with cross-attention to efficiently process large inputs across various modalities, while Bottleneck Transformers~\cite{nagrani2021attention, higuera2025tactile} introduce bottleneck tokens to explicitly restrict information flow within the network across various modalities. Motivated by these designs, M$\pi$NetsFusion further enhances this framework by employing a network architecture to adaptively attend to various planning sensing modalities for effective motion planning.
\section{\name} \label{sec: motiongeneralizer}
In this section, we introduce \name. It first creates a variety of common everyday articulated objects, as demonstrated in Figure~\ref{fig_2- motiongeneralizer-framework}. Then, it leverages an LLM~\cite{achiam2023gpt} to generate feasible workspace configurations. \name also includes sensing modality and manipulator-agnostic, scene-specific planning problem definition modules to enable data collection for various robotic manipulator embodiments.

\begin{figure*}[htbp] 
\centering
\includegraphics[width=0.9\textwidth]{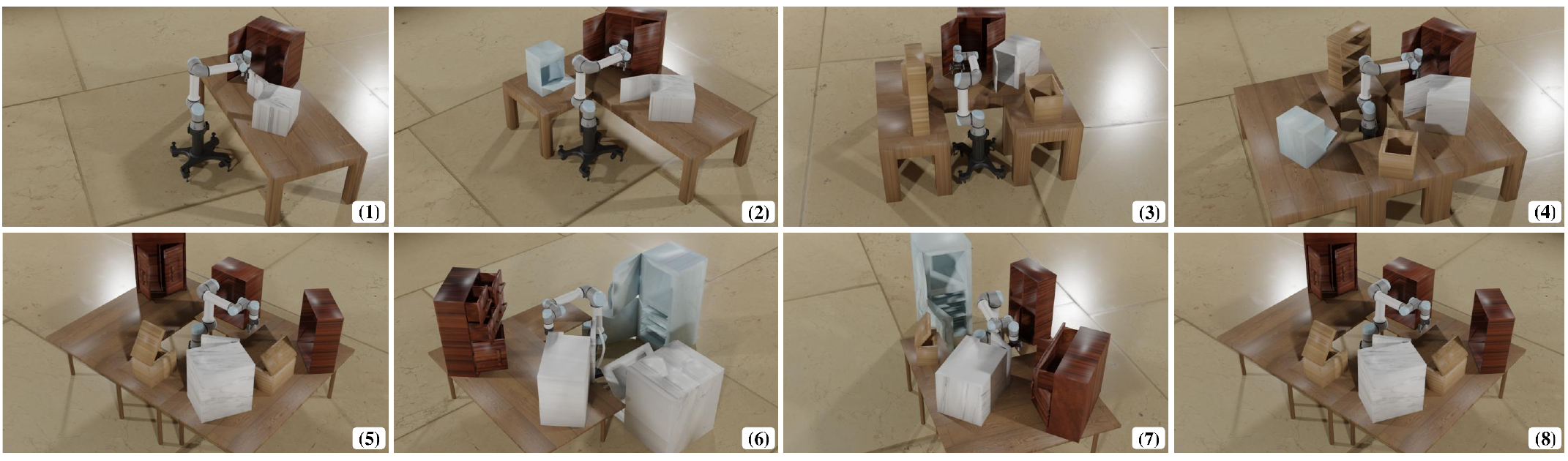}
\caption{\textbf{Workspace generation}: \name can generate an arbitrary number of cluttered, diverse workspaces given robotic manipulator type (UR5e) and a pool of everyday primitives.}
\label{fig_3- motiongeneralizer-nofeedback}
\end{figure*}

\begin{figure}[t] 
\centering
\includegraphics[width=3.0 in]{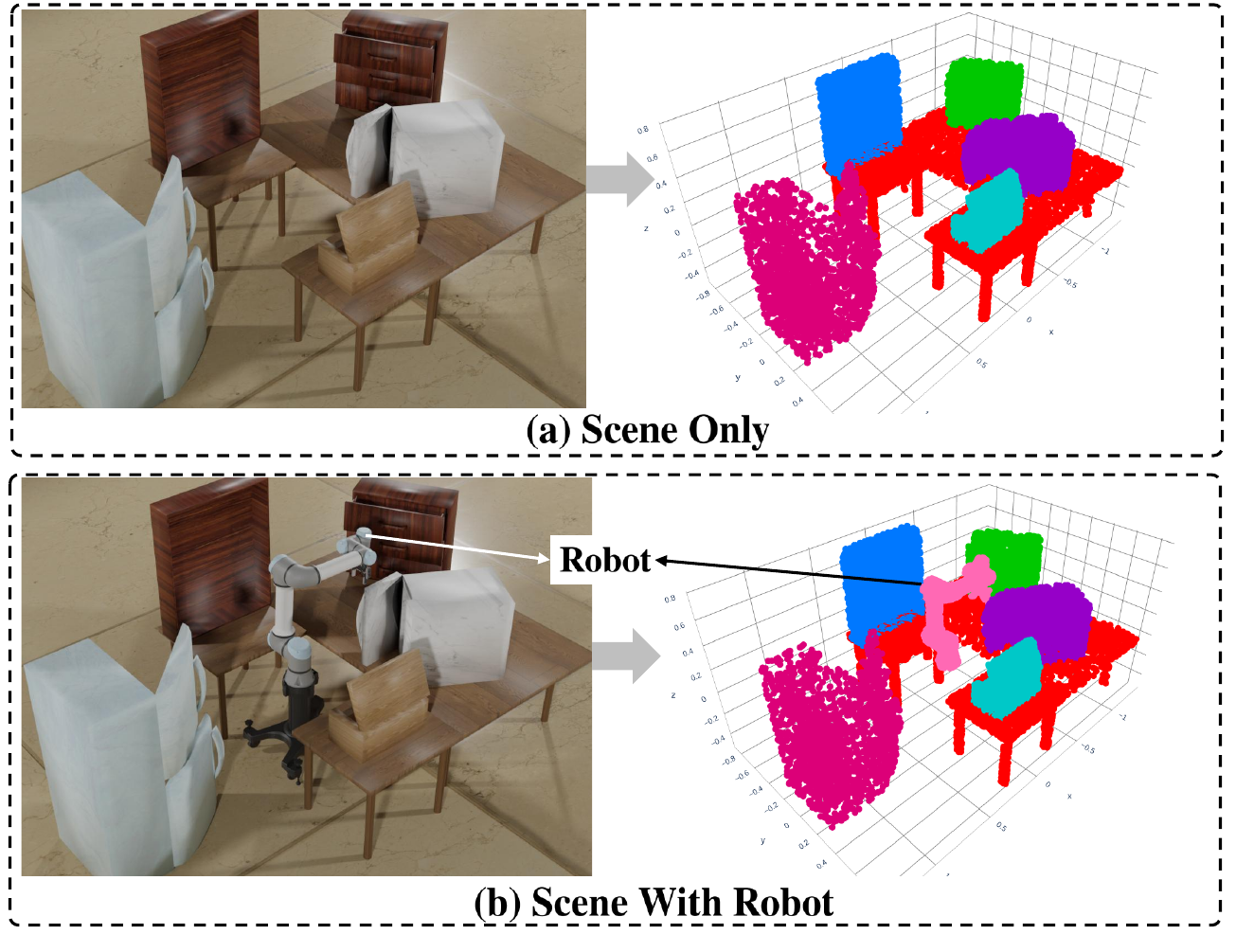}
\caption{\textbf{Point Cloud Synthesis}: \name's perception module provides a privileged point-cloud scene representation by uniformly sampling on workspace primitives (a) and robotic manipulator at arbitrary configurations (b) to increase spatial awareness for downstream planning tasks.}
\label{fig_4 - sensingmodality}
\end{figure}

\begin{figure*}[htbp] 
\centering
\centering
\includegraphics[width=\textwidth]{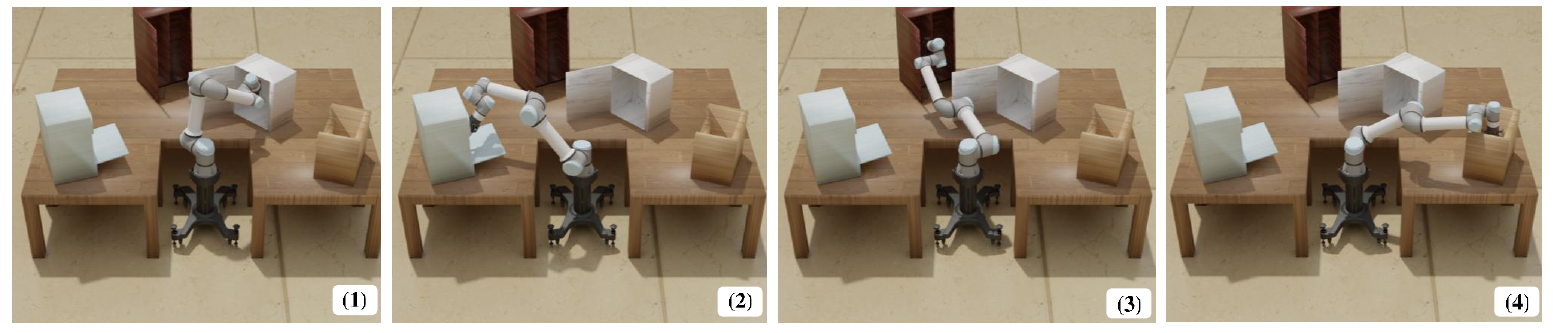}
\caption{\textbf{Planning Problem Generation}: \name's problem generator module provides robot-agnostic, scene-specific planning problem poses. Then, any arbitrary robotic manipulator, such as UR5e (top row), can use its collision-aware inverse kinematics to compute corresponding start-goal configurations of the planning problems.}
\label{fig_5- planningproblems}
\end{figure*}

\vspace{6pt}
\subsection{Procedural Asset Generation} \label{subsec: asset_generation}
For procedural asset generation, \name leverages methods from Neural MP \cite{dalal2024neural}, and NVIDIA's SceneSynthesizer \cite{eppner2025scene_synthesizer} framework. NeuralMP~\cite{dalal2024neural} leverages URDF sampling to generate customized URDFs for everyday primitives (e.g., tables, cabinets, dishwashers, microwaves, open boxes, and shelves) based on random ranges and numbers. \name also incorporates procedurally generated objects from SceneSynthesizer \cite{eppner2025scene_synthesizer}. These objects include active revolute and prismatic joints. To adapt them for motion planning, the active joints are fixed by assigning a random value within their original joint range.

\vspace{6pt}
\subsection{Language-guided Workspace Generation} \label{subsec: scenegen}
\noindent
\textbf{Manipulator and Table Selection.} \name first randomly selects a robotic manipulator with its reachability range, and randomly selects the number of tables to be placed around it. It then procedurally generates the required tables, using either Neural MP or the SceneSynthesizer framework, and places them within the manipulator’s reachability range.

\vspace{0.2cm}
\noindent
\textbf{Primitives per Table with LLM.} \name uses GPT-4 \cite{achiam2023gpt} to determine the number and types of primitives to place on each table. The prompt contains details such as the robotic manipulator’s specifications, the number and surface areas of the tables, and a list of common household objects. Then, the model outputs the quantity and types of primitives to be placed on each table. If fewer than four tables are generated, the LLM also suggests additional objects to place on the ground near the robot.

\vspace{0.2cm}
\noindent
\textbf{Procedural Primitive Generation.} Based on the types and quantities of primitives suggested by the LLM, \name procedurally generates each object using either the NeuralMP~\cite{dalal2024neural} or SceneSynthesizer~\cite{eppner2025scene_synthesizer} primitive pools. This randomized generation process ensures that objects of the same type vary in size and configuration, promoting diversity in the workspace.

\vspace{0.2cm}
\noindent 
\textbf{Primitive Placement with LLM.} Given the generated primitives on each table and on the ground, the LLM determines the pose of each object relative to the local coordinate frame of the corresponding table. The prompt to the language model includes the robot specifications, table surface areas, and the dimensions of each primitive.

\vspace{0.2cm}
\noindent
\textbf{Resolving Potential Collisions.} \name performs post-processing collision checking to detect and resolve any potential collisions in the workspace generated by the large language model. Figure \ref{fig_3- motiongeneralizer-nofeedback} provides some examples of the generated workspaces.

\vspace{6pt}
\subsection{Reducing Similarities Between Generated Workspaces} \label{subsec: rag}
\name already minimizes workspace similarity by randomly sampling the number of tables and generating individual primitives (Section \ref{subsec: scenegen}). \name introduces an additional framework that further reduces inter-workspace similarity, providing an extra layer of diversification.

This feedback framework leverages GPT-4~\cite{achiam2023gpt} to generate a high-level, distinctive description of each workspace. Based on the generated workspace description, the language model returns a concise textual description, which is then converted into an embedding vector utilizing OpenAI's \texttt{text-embedding3-large} model~\cite{achiam2023gpt}. This embedding is compared against those of previously generated workspaces to measure similarity. If the similarity is below a predefined threshold, the workspace is accepted.

\vspace{6pt}
\subsection{Perception Modality} \label{subsec: modality}
In real-world motion planning settings, the knowledge of precise geometries of obstacles are not known, and usually a sensed representation through RGB-D cameras may be available \cite{ramsey2024collision}. \name provides a high-dimensional point cloud workspace observation for training neural motion planners and performing collision checking.

Utilizing multiple RGB-D cameras in simulation for point cloud rendering would consume a substantial amount of time, and the resulting point clouds would be occluded. Instead, MotionGeneralizer generates a synthetic workspace point cloud by uniformly sampling points on the surface of objects within the workspace and robotic manipulator at arbitrary configurations (Figure \ref{fig_4 - sensingmodality}). This enables large-scale training of a privileged neural motion planner in simulation, which then can be fine-tuned with partial (occluded) camera-rendered point clouds before deployment in the real world \cite{lumdextrah}. Although training uses privileged point clouds, deployment relies on RGB-D-based reconstruction, which introduces a sim-to-real gap. However, this gap is less pronounced for motion planning than for manipulation, since the objective is collision avoidance, and sufficient camera coverage can provide an adequate reconstruction of the workspace.

\vspace{6pt}
\subsection{Motion Planning Problem Generator} \label{subsec: planningproblems}
\name generates robot-agnostic, scene-specific motion planning problems by sampling collision-free poses within the workspace, as demonstrated in Figure \ref{fig_5- planningproblems}. Firstly, it samples collision-free poses within the workspace primitives and queries the collision-aware inverse kinematics (IK) of the given robotic manipulator to determine the planning problem within the configuration space. Additionally, it samples collision-free poses within a hemisphere constructed based on the robotic manipulator's reachability radius. Given the motion planning problem, any off-the-shelf motion planner (i.e., motion planners from OMPL \cite{sucan2012open}, and cuRobo \cite{sundaralingam2023curobo}) can be utilized to plan and collect dataset for training neural motion planners.

\vspace{6pt}
\subsection{\name vs. \textsc{MotionBenchMaker}}
In this section, we qualitatively compare the workspaces generated by \textsc{MotionBenchMaker} \cite{chamzas2021motionbenchmaker} with those created by \name. 

\begin{figure*}[htbp] 
\centering
\includegraphics[width=0.9\textwidth]{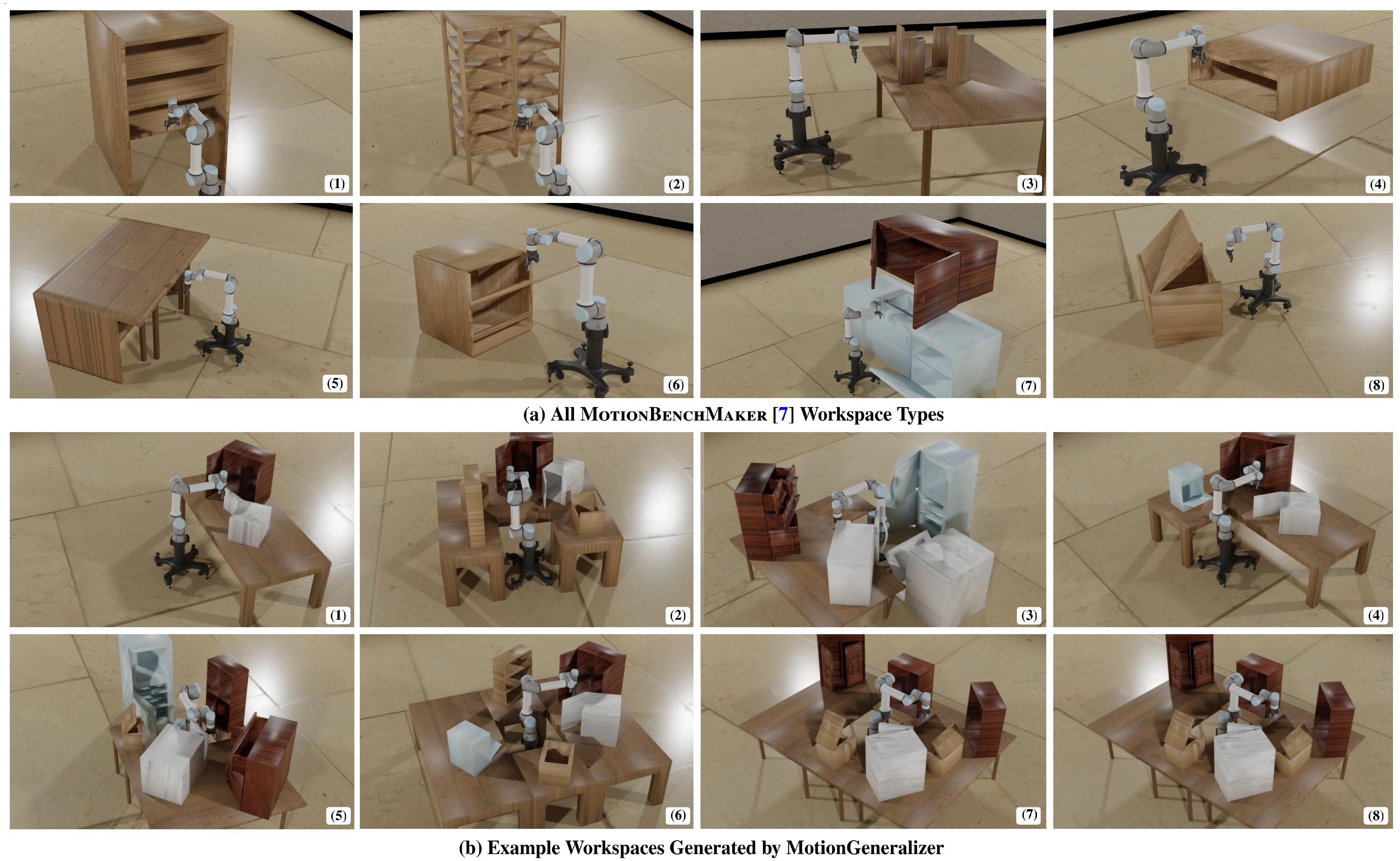}
\caption{\textbf{\name vs. \textsc{MotionBenchMaker} \cite{chamzas2021motionbenchmaker}}: \textsc{MotionBenchMaker} only contains eight different workspaces with one major obstacle present (top). \name can provide an arbitrary number of workspaces, and LLM determines the number of obstacles per scene (bottom - only 8 demonstrated for the sake of comparison), resulting in realistic and cluttered workspaces for robotic manipulator motion planning.}
\label{fig_6- motionbenchmaker}
\end{figure*}

As demonstrated in Figure \ref{fig_6- motionbenchmaker}, \textsc{MotionBenchMaker} generates simple workspaces containing only a single major obstacle. As a result, the generated workspaces are not realistic, and neural motion planners trained on these struggle to plan within real-world environments.

\vspace{6pt}
\subsection{Quantitative Evaluation of Workspace Generation}
We compare \name with a random workspace generation baseline~\cite{dalal2024neural} utilizing identical primitives to qualitatively validate the quality of LLM-guided workspace generation. We evaluate across the metrics relevant to motion planning. \textit{Scene collision rate} measures workspace feasibility by quantifying the percentage of scenes containing at least one object-object collision. \textit{Object collision rate} measures the frequency of invalid object placements by computing the proportion of objects involved in collisions. \textit{Object-category entropy} captures the diversity of object types within a workspace. \textit{Table surface coverage} measures scene density and clutter by estimating the proportion of occupied table area. The results are demonstrated in Table~\ref{tab:workspace_eval}.

\begin{table}[htbp]
\centering
\caption{Quantitative comparison of \name and random workspace generation baseline~\cite{dalal2024neural}.}
\label{tab:workspace_eval}
\resizebox{0.4\textwidth}{!}{
\begin{tabular}{lcc}
\hline
\textbf{Metric} & \textbf{Random} & \textbf{LLM-guided} \\
\hline
Scene Collision Rate (\%) & 53.00 & \textbf{14.00} \\
Object Collision Rate (\%) & 37.28 & \textbf{9.00} \\
Object Category Entropy & 0.74 & \textbf{0.92} \\
Table Surface Coverage (\%) & \textbf{20.26} & 14.62 \\
\hline
\end{tabular}}
\end{table}

The results demonstrate that \name improves the physical plausibility and structure of the generated workspaces. In particular, \name reduces the \textit{scene collision rate} from 53\% to 14\% and the \textit{object collision rate} from 37.28\% to 9\%, indicating that the proposed framework produces fewer physically infeasible workspace layouts. Furthermore, the higher \textit{object-category entropy} (0.92 vs. 0.74) suggests that LLM-generated workspaces contain more diverse object types. However, \name achieves lower \textit{table surface coverage} compared to the random workspace generation benchmark. Overall, these results show that MotionGeneralizer produces more physically feasible, diverse, and structured environments compared to random placement across the defined motion planning-specific metrics.

\vspace{8pt}
\section{M$\pi$NetsFusion} \label{sec: motionplanner}
In this section, we introduce M$\pi$NetsFusion, a neural motion planner that generates a feasible trajectory connecting given start and goal configurations in an end-to-end, open-loop manner.

\begin{figure*}[htbp] 
\centering
\includegraphics[width=0.9\textwidth]{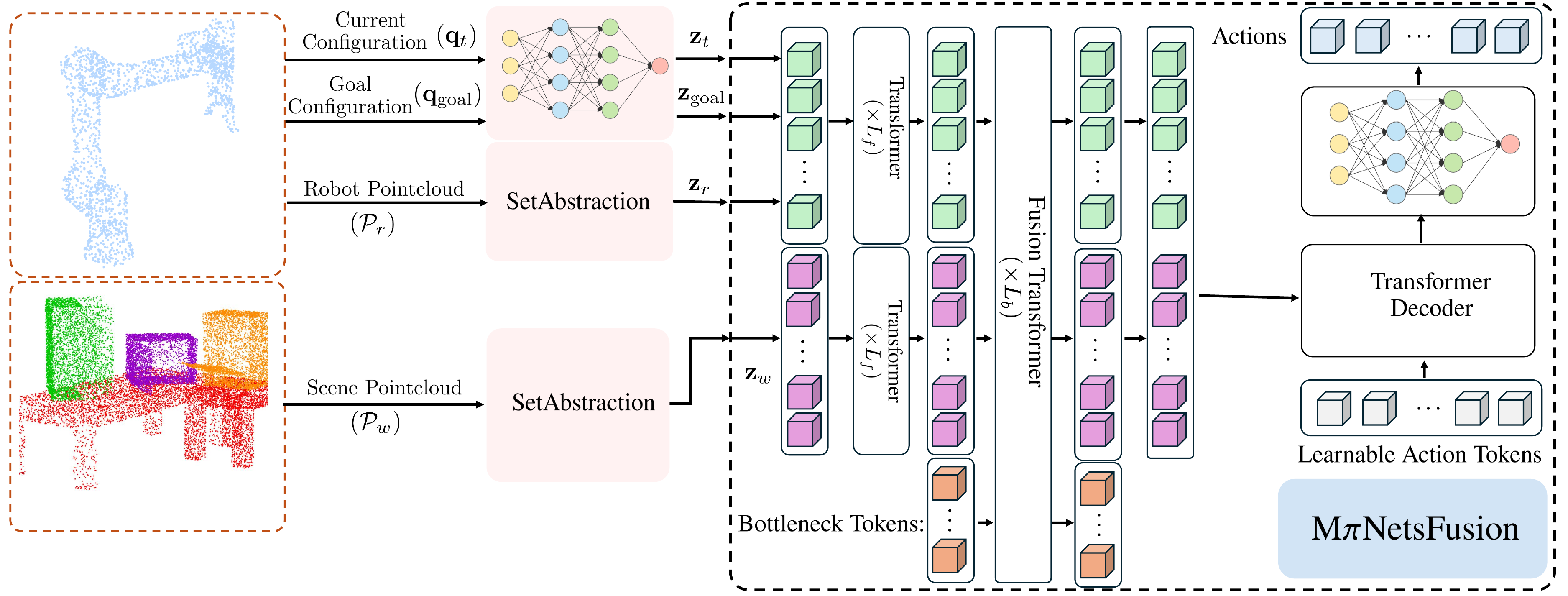}
\caption{\textbf{M$\pi$NetsFusion Network Architecture}: The framework leverages current-time-step workspace information (including the workspace and robot point clouds) together with configuration space features (current-time-step and goal configuration) to generate an action sequence that guides the robotic manipulator toward the motion planning goal. $\mathbf{z}_t$, $\mathbf{z}_{\text{goal}}$, $\mathbf{z}_r$, and $\mathbf{z}_w$ are the current time-step configuration, planning goal configuration, robot point cloud, and scene point cloud embeddings, respectively. $L_f$ is the number of unimodal transformer layers, and $L_b$ is the number of fusion transformer layers.}
\label{fig_7- network}
\end{figure*}

\vspace{6pt}
\subsection{Motion Planning Definition}
Let $\mathcal{C}\in \mathbb{R}^n$ denote the configuration space of a robotic manipulator, representing an $n$-dimensional space spanned by its joint values. This space consists of two regions: the obstacle space ($\mathcal{C}_{obs} \subset \mathcal{C}$), and the free space ($\mathcal{C}_{free}=\mathcal{C}\backslash \mathcal{C}_{obs}$). The objective of motion planning is to find a feasible path that connects the start and goal configurations while remaining entirely within the free space ($\mathcal{C}_{free}$).

\vspace{6pt}
\subsection{Planning via Action Chunking}
Inspired by the success of action chunking for robotic manipulation~\cite{zhao2023learning, chi2025diffusion} and motion planning~\cite{yang2025deep}, M$\pi$NetsFusion also utilizes an action chunking encoder-decoder framework for motion planning. Figure \ref{fig_7- network} demonstrates the structure of M$\pi$NetsFusion motion planner. 

\vspace{6pt}
\subsection{M$\pi$NetsFusion Components}
\noindent
\textbf{Embedding Current and Goal Configurations.} A shared multi-layer perceptron (MLP) tokenizes the current and goal configurations as follows:
\begin{equation} \label{eq - current}
\begin{aligned}
    \mathbf{z}_t = \text{MLP}_{I}(\mathbf{q}_t);~\mathbf{z}_{goal} = \text{MLP}_{I}(\mathbf{q}_{goal}),
\end{aligned}
\end{equation}
where $\mathbf{q}_t \in \mathbb{R}^6$, $\mathbf{q}_{goal} \in \mathbb{R}^6$, $\mathbf{z}_t \in \mathbb{R}^H$, and $\mathbf{z}_{goal} \in \mathbb{R}^H$ are current time-step configuration, goal configuration, current time-step configuration embedding, and goal configuration embedding, respectively.

\vspace{0.2cm}
\noindent
\textbf{Embedding Robot and Workspace Point Clouds.} The set abstraction layers from  PointNet++ \cite{qi2017pointnet++} are utilized to down-sample and tokenize these point clouds as follows:
\begin{equation} \label{eq - pcd_robot}
\begin{aligned}
    \mathbf{z}_r = \text{SetAbstraction}(\mathcal{P}_r);~\mathbf{z}_{w} = \text{SetAbstraction}(\mathcal{P}_w),
\end{aligned}
\end{equation}
where $\mathcal{P}_r \in \mathbb{R}^{N_r \times 3}$, $\mathcal{P}_w \in \mathbb{R}^{N_w \times 3}$, $\mathbf{z}_r \in \mathbb{R}^{K_r \times H}$, and $\mathbf{z}_w \in \mathbb{R}^{K_w \times H}$ are robot point cloud, scene point cloud, robot point cloud embedding, and scene point cloud embedding, respectively.

\vspace{0.2cm}
\noindent
\textbf{Transformer Encoder.} After encoding all planning modalities, we concatenate all robot-related modalities. Afterwards, we utilize positional encoding to retain the positional information of transformer encoder inputs~\cite{dosovitskiy2020image, higuera2025tactile}. After applying positional encoding, each planning modality can be represented as follows:
\begin{equation} \label{eq - current_embed}
\begin{aligned}
    \mathbf{z}_{\text{robot}} = [\mathbf{z}_{\text{robot}} + \mathbf{p}_{\text{robot}}];~\mathbf{z}_w = [\mathbf{z}_w + \mathbf{p}_w],
\end{aligned}
\end{equation}
where $\mathbf{p}_{\text{robot}} \in \mathbb{R}^{(K_r + 2) \times H}$,and $\mathbf{p}_w \in \mathbb{R}^{K_w \times H}$ are positional encodings for each planning sensing modalities.

Instead of performing pair-wise attention between planning sensing modalities, we introduce a set of fusion bottleneck tokens to the transformer encoder input sequence as follows:
\begin{equation} \label{eq - fusion}
\begin{aligned}
    \mathbf{z} = [\mathbf{z}_{\text{robot}} \parallel \mathbf{z}_w \parallel \mathbf{z}_{\text{fusion}}],
\end{aligned}
\end{equation}
where $\mathbf{z}_{\text{fusion}} = [\mathbf{z}_{\text{fusion}}^1, \cdots, \mathbf{z}_{\text{fusion}}^B] \in \mathbb{R}^{B \times H}$ is fusion bottleneck tokens. Then, the cross-attention between different planning sensing modalities is performed via these fusion bottleneck tokens as follows:
\begin{equation} \label{eq - fusion_encoder}
\begin{aligned}
    [\mathbf{z}_i^{l+1} \parallel \hat{\mathbf{z}}_{\text{fusion}_i}^{l+1}] = \text{TransformerEncoder}([\mathbf{z}_i^{l} \parallel \hat{\mathbf{z}}_{\text{fusion}_i}^{l}]; \mathbf{\theta_i}),
\end{aligned}
\end{equation}
\begin{equation} \label{eq - fusion_average}
\begin{aligned}
    \mathbf{z}_{\text{fusion}}^{l+1} = \frac{\sum_{i=1}^B \hat{\mathbf{z}}_{\text{fusion}_i}^{l+1}}{B},
\end{aligned}
\end{equation}
where $i$ and $\mathbf{\theta}_i$ are planning modality indexes, and network parameters, respectively. These bottleneck tokens distill and transform information between different planning sensing modalities within each transformer encoder layer. Then, we utilize a concatenation operator to aggregate the processed information from each modality, serving as the memory for the transformer decoder.

\begin{figure*}[htbp] 
\centering
\includegraphics[width=0.9\textwidth]{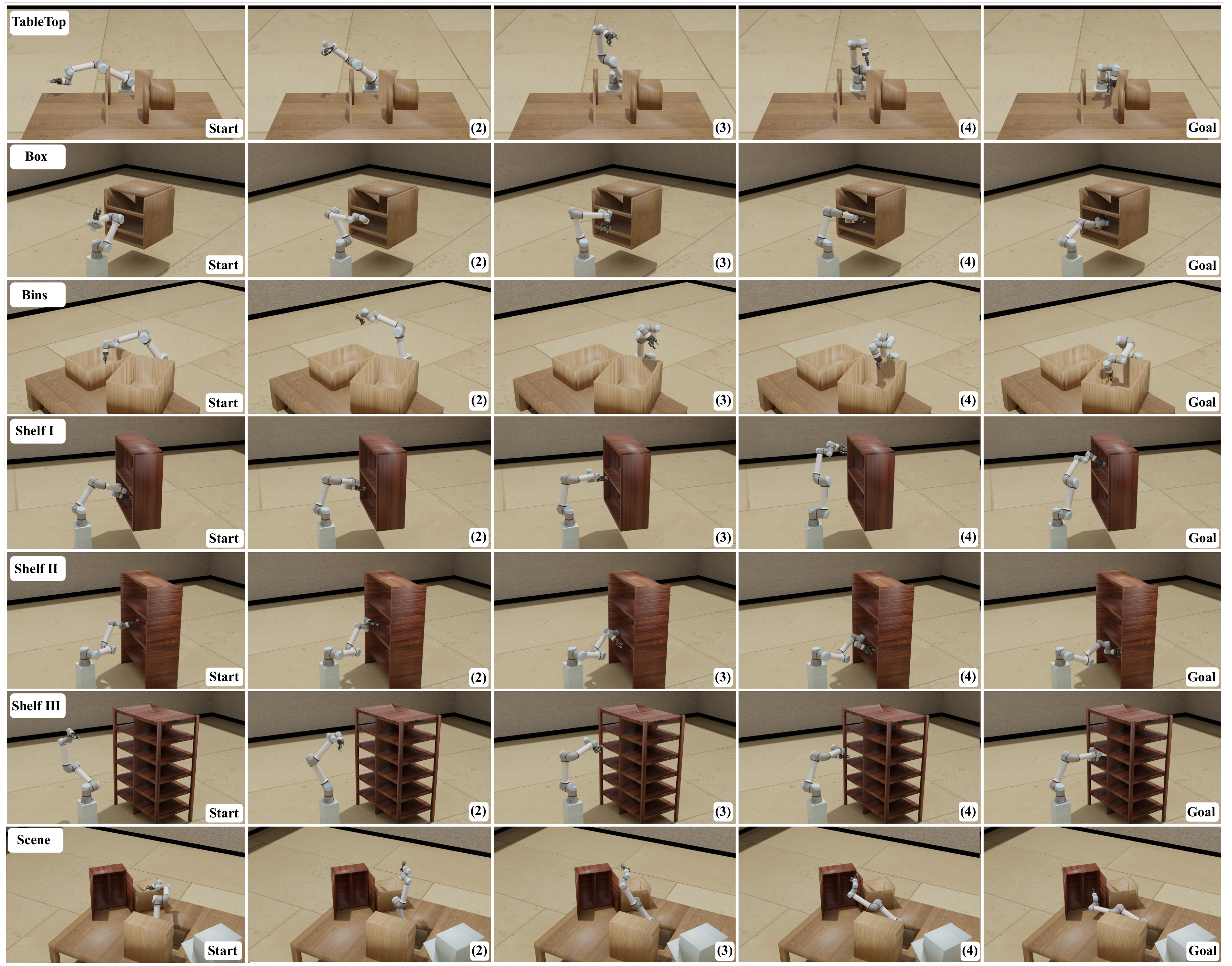}
\caption{\textbf{Simulation evaluation:} $\pi$NetsFusion planning in held-out planning task and an out-of-distribution \textbf{Scene} (bottom row) generated by \name with planning time of 0.2 $\sim$ 0.3 seconds. The path profile from start configuration to the goal configuration is demonstrated through frames 2-4.}
\label{fig_8- demo}
\end{figure*}

\vspace{0.2cm}
\noindent
\textbf{Transformer Decoder.} The transformer decoder follows a standard transformer architecture, and takes as input a set of learnable action tokens ($S \in \mathbb{R}^{A \times H}$) and uses the encoder output as memory to generate a sequence of delta joint actions $[\delta\mathbf{\bar{q}}_1, \delta\mathbf{\bar{q}}_2, \cdots, \delta\mathbf{\bar{q}}_A] \in \mathbb{R}^{A \times 6}$, which are then converted into joint targets and executed by the robotic manipulator’s low-level controller.

Table \ref{tab:mpinetsfusion_hparams} presents the architecture of M$\pi$NetsFusion. The design follows the concurrent architecture proposed in \cite{yang2025deep}, considering its demonstrated effectiveness for motion planning.

\begin{table}[t]
\centering
\caption{M$\pi$NetsFusion architecture and training hyperparameters. ``PCD'' denotes point cloud, and ``MHSA'' denotes multi-head self-attention.}
\label{tab:mpinetsfusion_hparams}
\renewcommand{\arraystretch}{1.12}
\setlength{\tabcolsep}{4pt}
\resizebox{0.45\textwidth}{!}{%
\begin{tabular}{@{}p{0.5\columnwidth}p{0.42\columnwidth}@{}}
\toprule
\textbf{Hyperparameter} & \textbf{Value} \\
\midrule
\multicolumn{2}{c}{\textbf{Input Parameters}} \\
\midrule
Scene-PCD ($|\mathcal{P}_w|$) & 2048 \\
Robot-PCD ($|\mathcal{P}_r|$) & 256 \\
\midrule
\multicolumn{2}{c}{\textbf{Training Parameters}} \\
\midrule
Optimizer & AdamW \\
Base Learning Rate & 0.001 \\
Weight Decay & 0.01 \\
Adam Momentum & $\beta_1{=}0.9,\ \beta_2{=}0.95$ \\
Batch Size & 256 \\
Learning-rate Schedule & cosine decay \\
Warmup Steps & 100 \\
Total Gradient Steps & 1.6M \\
Training/Validation Split & 90\% / 10\% \\
Validation Frequency & 10000 \\
GPU & NVIDIA A100 GPU (48gb) \\
Wall-Clock Time & 1 day \\
\midrule
\multicolumn{2}{c}{\textbf{PCD Set-Abstraction Layer}} \\
\midrule
Radius (m) & 0.1 \\
Samples Per Centroid & 64 \\
MLP Hidden Layers & [64, 64, 64, 128] \\
Scene-PCD Tokens ($K_w$) & 128 \\
Robot-PCD Tokens ($K_r$) & 16 \\
\midrule
\multicolumn{2}{c}{\textbf{Current/Target Joint Configuration Encoder}} \\
\midrule
MLP Hidden Layers & [128, 256, 128] \\
\midrule
\multicolumn{2}{c}{\textbf{TransformerFusion Parameters}} \\
\midrule
Unimodal Layers & 2 \\
Bottleneck Fusion Layers & 4 \\
Bottleneck Tokens & 4 \\
Decoder Layers & 6 \\
MHSA Heads & 8 \\
Hidden Dimension & 128 \\
Feed-Forward Dimension & 1024 \\
Positional Encoding & sin cos / learned \\
Action Chunk & 10 \\
\bottomrule
\end{tabular}}
\end{table}

\section{Results and Discussions} \label{sec:main_eval}
In this section, we detail the implementation of M$\pi$NetsFusion and compare its performance with state-of-the-art motion planners. All the frameworks are developed using the PyTorch framework \cite{paszke2017automatic}, and the simulation evaluations were conducted on a computer with NVIDIA RTX 4080 GPU. 

\vspace{6pt}
\subsection{Data Collection}
We leveraged \name (Section \ref{sec: motiongeneralizer}) to create a diverse set of planning workspaces, and motion planning problems for a UR5e robotic manipulator, and then leverage cuRobo \cite{sundaralingam2023curobo}, a GPU-accelerated optimization-based motion planner, to collect a dataset of $3.5$M trajectories for training the proposed neural motion planner over approximately 1000 workspaces generated via \name. The synthesis of the $3.5$M trajectory dataset required approximately six days of wall-clock time on an RTX A6000 GPU.

\vspace{6pt}
\subsection{M$\pi$NetsFusion Training}
We utilize M$\pi$NetsFusion parameterized by $\theta$ for end-to-end, open-loop planning. Let $[\delta\mathbf{q}_1, \delta\mathbf{q}_2, \cdots, \delta\mathbf{q}_A] \in \mathbb{R}^{A\times 6}$ be the ground-truth delta joint actions from the planning dataset $\mathcal{D}$, the training loss function is defined as:
\begin{equation} \label{eq - lossfn}
\begin{aligned}
    \mathcal{L}_{\text{M$\pi$NetsFusion}} = \frac{1}{A}\sum_{i=1}^A \parallel \delta\mathbf{q}_i - \delta\mathbf{\bar{q}}_i \parallel^2
\end{aligned}
\end{equation}
where $\delta\mathbf{\bar{q}}_i$ is the delta joint action predicted by the network. We train a $4.15$M parameter model on a dataset of $3.5$M trajectories. The network is optimized using an MSE loss for $1.6$M gradient steps, which requires a wall-clock time of one day of training on a single NVIDIA A100 GPU with a batch size of 256. Training hyperparameters are listed in Table~\ref{tab:mpinetsfusion_hparams}.

\vspace{6pt}
\subsection{M$\pi$NetsFusion Planning}
M$\pi$NetsFusion is evaluated in an open-loop manner, similar to the state-of-the-art end-to-end neural motion planners \cite{fishman2023motion, dalal2024local}. The planning policy is rolled out for a fixed number of steps, where each predicted joint action chunks is recursively fed back as input for subsequent action joint predictions. Planning success is defined as the end-effector reaching within a predefined threshold of the planning problem's goal-pose configuration while avoiding collisions with the environment. Algorithm \ref{alg:inference} summarizes motion planning with M$\pi$NetsFusion.

\begin{algorithm}[h]
\caption{M$\pi$NetsFusion Planning}
\label{alg:inference}
\DontPrintSemicolon

\KwIn{
Trained planner $\pi_\theta$; roll-out length $T$; action chunk size $a$; start and goal configurations $\mathbf{q}_{\mathrm{start}}, \mathbf{q}_{\mathrm{goal}}$; scene point cloud $\mathrm{ScenePCD}$; robot point-cloud sampler $\mathrm{PCDSampler}$; forward kinematics $\mathrm{FK}$; collision checker $Q$; success thresholds $\Delta_{\mathrm{pos}}, \Delta_{\mathrm{quat}}$.
}

\KwOut{
Collision-free trajectory $\tau$ or $\emptyset$.
}

\BlankLine

$\tau_{\mathrm{roll}} \leftarrow \{\mathbf{q}_{\mathrm{start}}\}$\;
$\mathbf{q}_t \leftarrow \mathbf{q}_{\mathrm{start}}$\;
$(\mathbf{p}_g, \mathbf{r}_g) \leftarrow \mathrm{FK}(\mathbf{q}_{\mathrm{goal}})$\;

\BlankLine
\tcp{Open-loop policy rollout}
\For{$i \leftarrow 1$ \KwTo $T$}{
    $\mathrm{RobotPCD} \leftarrow \mathrm{PCDSampler}(\mathbf{q}_t)$\;
    $\delta \mathbf{q}_{t:t+a} \leftarrow \pi_\theta(\mathbf{q}_t, \mathbf{q}_{\mathrm{goal}}, \mathrm{RobotPCD}, \mathrm{ScenePCD})$\;

    \For{each $\delta \mathbf{q}$ in $\delta \mathbf{q}_{t:t+a}$}{
        $\mathbf{q}_t \leftarrow \mathbf{q}_t + \delta \mathbf{q}$\;
        $\tau_{\mathrm{roll}} \leftarrow \tau_{\mathrm{roll}} \cup \{\mathbf{q}_t\}$\;
    }
}

\BlankLine
\tcp{Extract trajectory until goal is reached}
$\tau \leftarrow \emptyset$\;
\For{each $\mathbf{q}$ in $\tau_{\mathrm{roll}}$}{
    $\tau \leftarrow \tau \cup \{\mathbf{q}\}$\;
    $(\mathbf{p}, \mathbf{r}) \leftarrow \mathrm{FK}(\mathbf{q})$\;

    \If{$d(\mathbf{p}, \mathbf{p}_g) \leq \Delta_{\mathrm{pos}}$ \textbf{and} $d(\mathbf{r}, \mathbf{r}_g) \leq \Delta_{\mathrm{quat}}$}{
        \tcp{Verify final trajectory}
        \If{$Q(\tau) = \mathrm{CollisionFree}$}{
            \Return $\tau$\;
        }
        \Else{
            \Return $\emptyset$\;
        }
    }
}

\Return $\emptyset$\;
\end{algorithm}

\vspace{6pt}
\subsection{Baselines}
The baseline motion planners are as follows:

\begin{itemize}
    \item \textbf{Adaptively Informed Trees (AIT*)} \cite{strub2020adaptively} is an asymptotically optimal sampling-based motion planning algorithm that utilizes informed heuristic sampling to build a planning tree towards the goal. 
    
    \item \textbf{MPNets} \cite{qureshi2019motion} is a neural informed sampler that generates informed (goal-directed, collision-free) samples within the framework of sampling-based motion planners.
    
    \item \textbf{SIMPNet} \cite{soleymanzadeh2025simpnet} is a kinematics-aware neural sampler that generates informed samples towards the planning goal. This framework utilizes a cross-attention mechanism to condition the sampling network on workspace embeddings for sample generation.
\end{itemize}

\setlength{\tabcolsep}{2pt}
\begin{table*}[htbp]
\begin{center}
\captionof{table}{Planning performance of M$\pi$NetsFusion and baseline planners across all held-out planning task environments. ``T'' denotes \textit{planning time}, and ``S'' refers to success rate. $\downarrow$ indicates lower is better, and $\uparrow$ indicates higher is better.}

\label{tab: mpinetsfusion-performance}

\hspace{-0.2cm}
\renewcommand{\arraystretch}{1.1}
\resizebox{\textwidth}{!}{%
\begin{tabular*}{\linewidth}{@{}l
@{\extracolsep{\fill}}
c c c c c c c@{\extracolsep{\fill}} c
c c c c c c}
\toprule
\phantom{Var.} &  
\multicolumn{2}{c}{\textbf{TableTop}}&\multicolumn{2}{c}{\textbf{Box}}&\multicolumn{2}{c}{\textbf{Bins}} && \multicolumn{6}{c}{\textbf{Shelf}}\\
\cmidrule{2-3}
\cmidrule{4-5}
\cmidrule{6-7}
\cmidrule{9-14}
\phantom{Var.}&\multicolumn{6}{c}{\phantom{Var.}}&&\multicolumn{2}{c}{\textbf{Task I}}&\multicolumn{2}{c}{\textbf{Task II}}&\multicolumn{2}{c}{\textbf{Task III}} \\
\cmidrule{9-14}
& {T $[s]\downarrow$} & {S $[\%]\uparrow$} & {T $[s]\downarrow$} & {S $[\%]\uparrow$} & {T $[s]\downarrow$} & {S $[\%]\uparrow$} & {} & {T $[s]\downarrow$} & {S $[\%]\uparrow$}& {T $[s]\downarrow$} & {S $[\%]\uparrow$}& {T $[s]\downarrow$} & {S $[\%]\uparrow$} \\
\toprule
\textbf{AIT*} \cite{strub2020adaptively}&$1.02{\pm}0.0$&$31$\%&$1.02{\pm}0.0$&$57$\%&$1.02{\pm}0.0$&$69$\%&&$1.02{\pm}0.0$&$42$\%&$1.02{\pm}0.0$&$27$\%&$1.2{\pm}0.0$&$27$\% \\
\toprule
\textbf{MPNets} \cite{qureshi2019motion}&$5.79{\pm}9.2$&$49$\%&$2.8{\pm}1.69$&$67.3$\%&$3.63{\pm}1.25$&$84.2$\%&&$3.39{\pm}5.44$&$40$\%&$2.68{\pm}1.06$&$34$\%&$7{\pm}4.7$&$32$\% \\
\midrule
\textbf{SIMPNet} \cite{soleymanzadeh2025simpnet}&$4.54{\pm}5.32$&$\mathbf{67}$\%&$2.2{\pm}2.34$&$\mathbf{88.6}$\%&$3.55{\pm}1.71$&$\mathbf{95}$\%&&$2.21{\pm}1.25$&$\mathbf{44}$\%&$2.57{\pm}0.89$&$\mathbf{35}$\%&$3.44{\pm}1.19$&$33$\% \\
\toprule
\textbf{M$\pi$NetsFusion}&$\mathbf{0.22{\pm}0.01}$&$58$\%&$\mathbf{0.23{\pm}0.05}$&$61.3$\%&$\mathbf{0.24{\pm}0.04}$&$84.5$\%&&$\mathbf{0.24{\pm}0.03}$&$38.0$\%&$\mathbf{0.22{\pm}0.01}$&$34.3$\%&$\mathbf{0.22{\pm}0.01}$&$\mathbf{38.3}$\% \\
\bottomrule
\end{tabular*}}
\end{center}
\end{table*}

\begin{figure}[htbp] 
\centering
\includegraphics[width=0.45\textwidth]{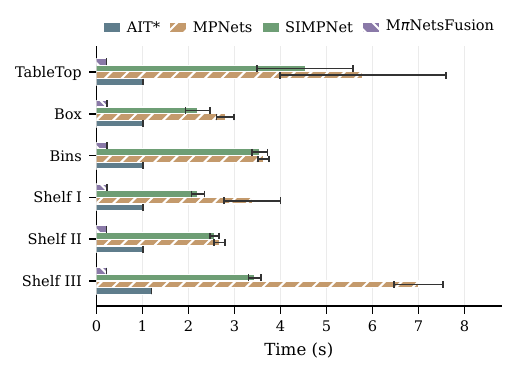}
\caption{\textbf{Planning-time Confidence Interval} of M$\pi$NetsFusion and all baselines across held-out planning tasks. Error bars indicate 95\% confidence intervals for planning time average computed for each task.}
\label{fig_9- confidence_interval}
\end{figure}

\begin{table}[htbp]
\centering
\caption{Average success rate and time comparison of M$\pi$NetsFusion with benchmark planners across all held-out planning tasks.}
\label{tab: average_success_rate}
\renewcommand{\arraystretch}{1.2}
\resizebox{0.48\textwidth}{!}{%
\begin{tabular}{@{}p{0.2\linewidth}p{0.13\linewidth}p{0.07\linewidth}p{0.08\linewidth}p{0.15\linewidth}p{0.2\linewidth}@{}}
\toprule
&&\multicolumn{1}{c}{\textbf{AIT*}}&\multicolumn{1}{c}{\textbf{MPNets}}&\multicolumn{1}{c}{\textbf{SIMPNet}}&\multicolumn{1}{c}{\textbf{M$\pi$NetsFusion}} \\
\toprule
\multicolumn{1}{l}{S $[\%]$} && \multicolumn{1}{c}{$42.1$\%} & \multicolumn{1}{c}{$51$\%} & \multicolumn{1}{c}{$51.8$\%}& \multicolumn{1}{c}{$52.4$\%} \\
\toprule
\multirow{6}{*}{$\frac{\text{T}}{\text{T}_{\text{M$\pi$NetsFusion}}}$}&\textbf{TableTop}&\multicolumn{1}{c}{$4.63$}&\multicolumn{1}{c}{$26.31$}&\multicolumn{1}{c}{$20.5$}&\multicolumn{1}{c}{$1.0$} \\ \hhline{~-----}
                                                              &\textbf{Box}&\multicolumn{1}{c}{$4.43$}&\multicolumn{1}{c}{$12.17$}&\multicolumn{1}{c}{$9.56$}&\multicolumn{1}{c}{$1.0$} \\  \hhline{~-----}
                                                              &\textbf{Bins}&\multicolumn{1}{c}{$4.25$}&\multicolumn{1}{c}{$15.12$}&\multicolumn{1}{c}{$14.79$}&\multicolumn{1}{c}{$1.0$} \\  \hhline{~-----}
                                                              &\textbf{Shelf I}&\multicolumn{1}{c}{$4.25$}&\multicolumn{1}{c}{$14.12$}&\multicolumn{1}{c}{$9.20$}&\multicolumn{1}{c}{$1.0$} \\  \hhline{~-----}
                                                              &\textbf{Shelf II}&\multicolumn{1}{c}{$4.63$}&\multicolumn{1}{c}{$12.18$}&\multicolumn{1}{c}{$11.68$}&\multicolumn{1}{c}{$1.0$} \\  \hhline{~-----}
                                                              &\textbf{Shelf III}&\multicolumn{1}{c}{$4.63$}&\multicolumn{1}{c}{$31.81$}&\multicolumn{1}{c}{$15.63$}&\multicolumn{1}{c}{$1.0$} \\ \hhline{~-----}
                                                              &\textbf{Average}&\multicolumn{1}{c}{$4.47$}&\multicolumn{1}{c}{$18.61$}&\multicolumn{1}{c}{$13.56$}&\multicolumn{1}{c}{$1.0$} \\
\toprule
\end{tabular}}
\end{table}

\setlength{\tabcolsep}{2pt}
\begin{table*}[htbp]
\begin{center}
\captionof{table}{Planning performance of M$\pi$NetsFusion and baseline planners within an out-of-distribution environment (\textbf{Scene}) generated by \name. ``T'' denotes \textit{planning time}, `N$_r$'' denotes percentage of planning problems reached the goal, and ``S'' refers to success rate. $\downarrow$ indicates lower is better, and $\uparrow$ indicates higher is better.}

\label{tab: mpinetsfusion-performance-outofdist}

\hspace{-0.2cm}
\renewcommand{\arraystretch}{1.1}
\resizebox{\textwidth}{!}{%
\begin{tabular*}{\linewidth}{@{}l
@{\extracolsep{\fill}}
c c c c c c c@{\extracolsep{\fill}} c
c c c c c c}
\toprule
\phantom{Var.} & \multicolumn{2}{c}{\textbf{AIT*} \cite{strub2020adaptively}}&\multicolumn{2}{c}{\textbf{MPNet} \cite{qureshi2019motion}}&\multicolumn{2}{c}{\textbf{SIMPNet} \cite{soleymanzadeh2025simpnet}} && \multicolumn{3}{c}{\textbf{NeuralMP} \cite{dalal2024neural}}&\multicolumn{3}{c}{\textbf{M$\pi$NetsFusion}}\\
\cmidrule{2-3}
\cmidrule{4-5}
\cmidrule{6-7}
\cmidrule{9-11}
\cmidrule{12-14}
& {T $[s]\downarrow$} & {S $[\%]\uparrow$} & {T $[s]\downarrow$} & {S $[\%]\uparrow$} & {T $[s]\downarrow$} & {S $[\%]\uparrow$} & {} & {T $[s]\downarrow$} & {N$_r$ $[\%]\uparrow$}& {S $[\%]\uparrow$} & {T $[s]\downarrow$}& {N$_r$ $[\%]\uparrow$} & {S $[\%]\uparrow$} \\
\toprule
\textbf{Scene}&$1.02 {\pm} 0.0$&$31$\%&$5.81 {\pm} 4.30$&$36$\%&$7.21 {\pm} 10.58$&$\mathbf{38}$\%&&$0.87 {\pm} 0.0$&$1.0$\%&$1.0$\%&$\mathbf{0.28 {\pm}0.02}$&$\mathbf{100}$\%&$28$\% \\
\toprule
\end{tabular*}}
\end{center}
\end{table*}

\vspace{6pt}
\subsection{Evaluation Metrics}
We use three planning metrics throughout the paper to evaluate the performance of M$\pi$NetsFusion against benchmark planners: \textit{planning time}, \textit{goal reaching}, and \textit{success rate}. \textit{Planning time} ``$T$'' denotes the average planning time the planner takes in each workspace. \textit{Goal reaching} ``$N_r$'' refers to the percentage of planned paths reached the goal in each environment, regardless of collision status, which is used to measure the performance of benchmark and ablation end-to-end open-loop neural motion planners. \textit{Success rate} ``$S$'' represents the percentage of successfully planned paths in each workspace.

\vspace{6pt}
\subsection{Evaluation Tasks}
We created several held-out planning environments for performance evaluation. These evaluation tasks are as follows:

\begin{itemize}
    \item \textbf{TableTop}: Adapted from \textsc{MotionBenchMaker} \cite{chamzas2021motionbenchmaker}, which consists of a tabletop environment with various obstacles placed on it in front of the robot.
    
    \item \textbf{Box}: Adapted from \textsc{MotionBenchMaker} \cite{chamzas2021motionbenchmaker}, which consists of a box placed in front of the robotic manipulator at arbitrary positions and orientations.
    
    \item \textbf{Bins}: Adapted from NeuralMP~\cite{dalal2024neural} which consists of two bins placed in front of the robot at arbitrary positions and orientations.
    
    \item \textbf{Shelf I, II \& III}: Adapted from \textsc{MotionBenchMaker} \cite{chamzas2021motionbenchmaker} and Neural MP \cite{dalal2024neural}, which consist of shelves placed in front of the robot at arbitrary configurations.
\end{itemize}

\vspace{6pt}
\subsection{Evaluation Results}
Table \ref{tab: mpinetsfusion-performance} and Figure~\ref{fig_9- confidence_interval} show the performance of the considered planners across these held-out planning environments, while Table \ref{tab: average_success_rate} provides the average success rate and time comparison of the proposed planner with benchmark planners across all held-out planning tasks. Also, Figure \ref{fig_8- demo} provides an example of policy roll-out for all the evaluation tasks in simulation.

M$\pi$NetsFusion achieves comparable performance to AIT* ($52.4$\% vs. $42.1$\%) while plan $4.47$ times faster on average. This improvement can be attributed to the fact that AIT* requires a collision checker to plan and iteratively check the collision of sampled states, whereas M$\pi$NetsFusion plans end-to-end, using only the planning start configuration, goal configuration, and the workspace point cloud.

M$\pi$NetsFusion also performs comparably to benchmark neural samplers. Its achieves a success rate of $52.4$\% versus $51$\% for MPNets and $51.8$\% for SIMPNet, while planing $18.61$ and $13.56$ times faster, respectively. MPNets flattens and concatenates configuration-space and workspace point cloud embeddings to predict the next configuration state towards the planning goal. Its learned sampling primitive is embedded within a closed-loop, bidirectional planner that falls back to replanning when two consecutive states cannot be connected. Consequently, MPNets relies on a collision-checker for both planning and replanning, resulting in larger planning times. SIMPNet leverages the kinematic structure of the robotic manipulator and employs a cross-attention mechanism to combine planning modalities from workspace and configuration space to predict the next time step configuration state towards the goal. However, similar to MPNets, SIMPNet is also embedded within a bi-directional, closed-loop planner that requires a collision checker, resulting in longer planning times.

We also generate an out-of-distribution workspace using \name to further evaluate the performance of M$\pi$NetsFusion against benchmark planners, as shown in Figure~\ref{fig_8- demo} and summarized in Table~\ref{tab: mpinetsfusion-performance-outofdist}. The planning results in this workspace follow the same performance trends observed in the held-out planning tasks.

Overall, the average success rate of M$\pi$NetsFusion, and all benchmark planners, is primarily influenced by lower performance in shelf environments, which can be attributed to the limited representation of shelf-like primitives in the training dataset and the inherent difficulty of narrow-passage scenarios. Shelf environments typically involve tight spaces, which are challenging for open-loop end-to-end planning. Since M$\pi$NetsFusion does not perform iterative collision checking during inference, it is more sensitive to small deviations that can lead to collisions in these tight planning environments.

\subsection{Comparison with NeuralMP}
\begin{figure*}[t] 
\centering
\includegraphics[width=0.9\textwidth]{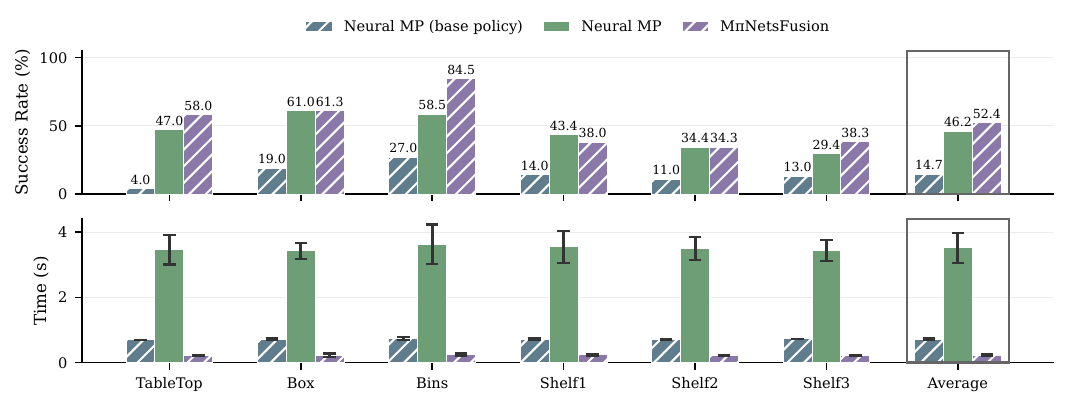}
\caption{\textbf{M$\pi$NetsFusion and NeuralMP~\cite{dalal2024neural}} performance comparison across all held-out planning task environments.}
\label{fig_9- neuralmp_comparison}
\end{figure*}

Neural MP~\cite{dalal2024neural} is an end-to-end neural motion planner that plans in an open-loop manner. This framework utilizes a Recurrent Neural Network (RNN)-based decoder to learn the temporal dependencies over the motion planning context, as well as a Gaussian Mixture Model (GMM) head to encode and mimic the probabilistic and multi-modal nature of sampling-based motion planning algorithms. We adopt Neural MP (base policy and policy with GMM head) based on the official implementation to our UR5e robotic manipulator. The planning results are depicted in Figure~\ref{fig_9- neuralmp_comparison}.

The base policy operates as an open-loop planner without multi-modal sampling, and its direct concatenation of configuration-space and workspace features can lead to dominance of workspace pointcloud embedding. In contrast, M$\pi$NetsFusion leverages a bottleneck fusion architecture to attend intelligently to various motion planning modalities, which yields higher success rates. Additionally, compared with NeuralMP with GMM head, M$\pi$NetsFusion yields faster planning while maintaining similar performance in terms of success rate.

\vspace{6pt}
\subsection{Comparison with Diffusion Policy}
\begin{figure}[htbp] 
\centering
\includegraphics[width=0.48\textwidth]{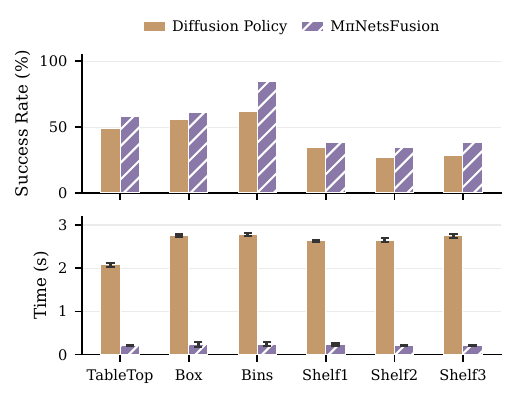}
\caption{\textbf{Diffusion Policy~\cite{chi2025diffusion} vs. M$\pi$NetsFusion} performance across held-out planning evaluation tasks.}
\label{fig_11- diffusion_policy}
\end{figure}

We compare the performance of M$\pi$NetsFusion with Diffusion Policy~\cite{chi2025diffusion} across the planning tasks. To enable this comparison, we replace the encoder in the original Diffusion Policy with the ACT transformer encoder (Figure~\ref{fig_9 - act}). The model is trained and evaluated using 100 diffusion steps and deployed in an end-to-end, closed-loop manner, consistent with M$\pi$NetsFusion. The results are shown in Figure~\ref{fig_11- diffusion_policy}.

The results show that M$\pi$NetsFusion achieves comparable success rates to the Diffusion Policy, while requiring lower average planning time. This is because the Diffusion Policy relies on 100 denoising steps, which increases inference time during end-to-end, open-loop planning.

\subsection{Failure Analysis}
The results in Table~\ref{tab: mpinetsfusion-performance} show that M$\pi$NetsFusion and all benchmark neural planners achieve low performance on shelf planning tasks. To further analyze this behavior, we added additional shelf environments with various articulations and provided a more detailed evaluation across evaluation tasks. We report self-collision rates to assess trajectory quality and scene-collision rates to capture interactions in narrow shelf spaces. The results are summarized in Table~\ref{tab: shelffailureanalysis}.

\begin{table}[htbp]
\centering
\caption{M$\pi$NetsFusion failure analysis across held-out shelf evaluation environments.}
\label{tab: shelffailureanalysis}
\resizebox{0.49\textwidth}{!}{%
\begin{tabular}{@{}p{0.20\linewidth}p{0.10\linewidth}p{0.1\linewidth}p{0.1\linewidth}p{0.1\linewidth}p{0.1\linewidth}p{0.1\linewidth}@{}}
\toprule
\multicolumn{1}{l}{\phantom{Var.}}&\multicolumn{1}{c}{\scriptsize{\textbf{Reached}}}&\multicolumn{1}{c}{\scriptsize{\textbf{Success}}}&\multicolumn{1}{c}{\scriptsize{\textbf{Collision}}}&\multicolumn{1}{c}{\scriptsize{\textbf{Self-Collision}}}&\multicolumn{1}{c}{\scriptsize{\textbf{Scene-Collision}}}&\multicolumn{1}{c}{\scriptsize{\textbf{Self\&Scene-Collision}}} \\
\toprule
\multicolumn{1}{l}{\scriptsize{\textbf{TableTop}}}&\multicolumn{1}{c}{\scriptsize{100\%}}&\multicolumn{1}{c}{\scriptsize{58\%}}&\multicolumn{1}{c}{\scriptsize{42\%}}&\multicolumn{1}{c}{\scriptsize{4\%}}&\multicolumn{1}{c}{\scriptsize{\underline{32}\%}}&\multicolumn{1}{c}{\scriptsize{6\%}} \\
\toprule
\multicolumn{1}{l}{\scriptsize{\textbf{Box}}}&\multicolumn{1}{c}{\scriptsize{100\%}}&\multicolumn{1}{c}{\scriptsize{61.3\%}}&\multicolumn{1}{c}{\scriptsize{38.7\%}}&\multicolumn{1}{c}{\scriptsize{10.7\%}}&\multicolumn{1}{c}{\scriptsize{\underline{21}\%}}&\multicolumn{1}{c}{\scriptsize{7\%}} \\
\toprule
\multicolumn{1}{l}{\scriptsize{\textbf{Bins}}}&\multicolumn{1}{c}{\scriptsize{100\%}}&\multicolumn{1}{c}{\scriptsize{84.5\%}}&\multicolumn{1}{c}{\scriptsize{15.5\%}}&\multicolumn{1}{c}{\scriptsize{2.0\%}}&\multicolumn{1}{c}{\scriptsize{\underline{12.5}\%}}&\multicolumn{1}{c}{\scriptsize{1.0\%}} \\
\toprule
\multicolumn{1}{l}{\scriptsize{\textbf{Shelf1}}}&\multicolumn{1}{c}{\scriptsize{100\%}}&\multicolumn{1}{c}{\scriptsize{38.0\%}}&\multicolumn{1}{c}{\scriptsize{62\%}}&\multicolumn{1}{c}{\scriptsize{9.7\%}}&\multicolumn{1}{c}{\scriptsize{\underline{48.7}\%}}&\multicolumn{1}{c}{\scriptsize{3.6\%}} \\
\toprule
\multicolumn{1}{l}{\scriptsize{\textbf{Shelf2}}}&\multicolumn{1}{c}{\scriptsize{100\%}}&\multicolumn{1}{c}{\scriptsize{34.3\%}}&\multicolumn{1}{c}{\scriptsize{65.7\%}}&\multicolumn{1}{c}{\scriptsize{12.7\%}}&\multicolumn{1}{c}{\scriptsize{\underline{48.0}\%}}&\multicolumn{1}{c}{\scriptsize{5.0\%}} \\
\toprule
\multicolumn{1}{l}{\scriptsize{\textbf{Shelf3}}}&\multicolumn{1}{c}{\scriptsize{97.3\%}}&\multicolumn{1}{c}{\scriptsize{38.3\%}}&\multicolumn{1}{c}{\scriptsize{59\%}}&\multicolumn{1}{c}{\scriptsize{10.3\%}}&\multicolumn{1}{c}{\scriptsize{\underline{44.7}\%}}&\multicolumn{1}{c}{\scriptsize{4\%}} \\
\toprule
\multicolumn{1}{l}{\scriptsize{\textbf{Shelf4}}}&\multicolumn{1}{c}{\scriptsize{100\%}}&\multicolumn{1}{c}{\scriptsize{34.3\%}}&\multicolumn{1}{c}{\scriptsize{65.7\%}}&\multicolumn{1}{c}{\scriptsize{7.7\%}}&\multicolumn{1}{c}{\scriptsize{\underline{52.3}\%}}&\multicolumn{1}{c}{\scriptsize{5.7\%}} \\
\toprule
\multicolumn{1}{l}{\scriptsize{\textbf{Shelf5}}}&\multicolumn{1}{c}{\scriptsize{100\%}}&\multicolumn{1}{c}{\scriptsize{28.3\%}}&\multicolumn{1}{c}{\scriptsize{71.7\%}}&\multicolumn{1}{c}{\scriptsize{9.3\%}}&\multicolumn{1}{c}{\scriptsize{\underline{50.0}\%}}&\multicolumn{1}{c}{\scriptsize{12.3\%}} \\
\toprule
\multicolumn{1}{l}{\scriptsize{\textbf{Shelf6}}}&\multicolumn{1}{c}{\scriptsize{99.7\%}}&\multicolumn{1}{c}{\scriptsize{25.3\%}}&\multicolumn{1}{c}{\scriptsize{74.3\%}}&\multicolumn{1}{c}{\scriptsize{13.0\%}}&\multicolumn{1}{c}{\scriptsize{\underline{51.7}\%}}&\multicolumn{1}{c}{\scriptsize{9.7\%}} \\
\bottomrule
\end{tabular}}
\end{table}

The results show that most failures in shelf environments are due to collisions with the environment. This high collision rate is likely due to the limited representation of shelf-like primitives in the \name primitive pool and their tight space configuration.

\vspace{6pt}
\subsection{Ablation Studies} \label{subsec: ablations}
In this section, we compare the performance of M$\pi$NetsFusion against ablations of its components and architecture to evaluate the contribution and impact of each module on overall performance.

\setlength{\tabcolsep}{2pt}
\begin{table*}[htbp]
\begin{center}
\captionof{table}{Planning comparison between M$\pi$NetsFusion and different frameworks that differently combine planning sensing modalities. ``N$_r$'' denotes percentage of planning problems reached the goal, and ``S'' denotes \textit{success rate}. $\uparrow$ indicates higher is better. ``M$\pi$NetsFusion-All'' denotes the framework that planning Embeddings Considered Separately, and ``M$\pi$NetsFusion-Config'' denotes the config that configuration space embeddings are combined.}

\label{tab: ablation-frameworks}

\hspace{-0.2cm}
\resizebox{\textwidth}{!}{%
\begin{tabular*}{\linewidth}{@{}l
@{\extracolsep{\fill}}
c c c c c c c@{\extracolsep{\fill}} c
c c c c c c}
\toprule
\phantom{Var.} &  
\multicolumn{2}{c}{\textbf{TableTop}}&\multicolumn{2}{c}{\textbf{Box}}&\multicolumn{2}{c}{\textbf{Bins}} && \multicolumn{6}{c}{\textbf{Shelf}}\\
\cmidrule{2-3}
\cmidrule{4-5}
\cmidrule{6-7}
\cmidrule{9-14}
\phantom{Var.}&\multicolumn{6}{c}{\phantom{Var.}}&&\multicolumn{2}{c}{\textbf{Task I}}&\multicolumn{2}{c}{\textbf{Task II}}&\multicolumn{2}{c}{\textbf{Task III}} \\
\cmidrule{9-14}
& {N$_r$ $[\%]\uparrow$} & {S $[\%]\uparrow$} & {N$_r$ $[\%]\uparrow$} & {S $[\%]\uparrow$} & {N$_r$ $[\%]\uparrow$} & {S $[\%]\uparrow$} & {} & {N$_r$ $[\%]\uparrow$} & {S $[\%]\uparrow$}& {N$_r$ $[\%]\uparrow$} & {S $[\%]\uparrow$}& {N$_r$ $[\%]\uparrow$} & {S $[\%]\uparrow$} \\
\toprule
\textbf{ACT} \cite{yang2025deep, zhao2023learning}&$87$\%&$37$\%&$86$\%&$51$\%&$88.5$\%&$65.25$\%&&$69.66$\%&$30.6$\%&$74$\%&$27$\%&$85.3$\%&$28.3$\% \\
\toprule
\textbf{M$\pi$NetsFusion-All}&$96$\%&$49$\%&$97.6$\%&$61$\%&$\mathbf{100}$\%&$\mathbf{86.75}$\%&&$\mathbf{100}$\%&$34$\%&$95$\%&$31$\%&$93$\%&$36.6$\% \\
\midrule
\textbf{M$\pi$NetsFusion-Config}&$\mathbf{100}$\%&$57$\%&$99.3$\%&$60.3$\%&$\mathbf{100}$\%&$85$\%&&$\mathbf{100}$\%&$\mathbf{40.3}$\%&$98.3$\%&$34$\%&$93.3$\%&$\mathbf{40.6}$\% \\
\midrule
\textbf{ViTacFormer} \cite{heng2025vitacformer}&$\mathbf{100}$\%&$57$\%&$\mathbf{100}$\%&$\mathbf{62}$\%&$\mathbf{100}$\%&$85.75$\%&&$\mathbf{100}$\%&$35.7$\%&$98.3$\%&$\mathbf{34.6}$\%&$\mathbf{99.33}$\%&$33.4$\% \\
\toprule
\textbf{M$\pi$NetsFusion}&$\mathbf{100}$\%&$58$\%&$\mathbf{100}$\%&$61.3$\%&$\mathbf{100}$\%&$84.5$\%&&$\mathbf{100}$\%&$38.0$\%&$\mathbf{100}$\%&$34.3$\%&$97.3$\%&$38.3$\% \\
\bottomrule
\end{tabular*}}
\end{center}
\end{table*}

\begin{table}[htbp]
\centering
\caption{Network parameter size and cold-start time comparison between M$\pi$NetsFusion and ablative structures.}
\label{tab: coldstart_time}
\begin{tabular}{@{}p{0.35\linewidth}p{0.3\linewidth}p{0.2\linewidth}@{}}
\toprule
&Cold Start [$ms$] $\downarrow$& Network Size \\
\toprule
\textbf{ACT} \cite{yang2025deep, zhao2023learning}&\multicolumn{1}{c}{$\mathbf{3.8}$}&\multicolumn{1}{c}{$\mathbf{4.52}$M} \\
\midrule
\textbf{M$\pi$NetsFusion-All}&\multicolumn{1}{c}{$7$}&\multicolumn{1}{c}{$7.32$M} \\
\midrule
\textbf{M$\pi$NetsFusion-Config}&\multicolumn{1}{c}{$4.6$}&\multicolumn{1}{c}{$6.1$M} \\
\midrule
\textbf{ViTacFormer} \cite{heng2025vitacformer}&\multicolumn{1}{c}{$5.01$}&\multicolumn{1}{c}{$4.6$M} \\
\midrule
\textbf{M$\pi$NetsFusion}&\multicolumn{1}{c}{$4.1$}&\multicolumn{1}{c}{$4.9$M} \\
\bottomrule
\end{tabular}
\end{table}

\setlength{\tabcolsep}{2pt}
\begin{table*}[htbp]
\begin{center}
\captionof{table}{Planning comparison between M$\pi$NetsFusion and ablation frameworks with different action chunk sizes in terms of planning time and success rate.}

\label{tab: action_chunks}

\hspace{-0.2cm}
\resizebox{\textwidth}{!}{%
\begin{tabular*}{\linewidth}{@{}l
@{\extracolsep{\fill}}
c c c c c c c@{\extracolsep{\fill}} c
c c c c c c}
\toprule
\phantom{Var.} &  
\multicolumn{2}{c}{\textbf{TableTop}}&\multicolumn{2}{c}{\textbf{Box}}&\multicolumn{2}{c}{\textbf{Bins}} && \multicolumn{6}{c}{\textbf{Shelf}}\\
\cmidrule{2-3}
\cmidrule{4-5}
\cmidrule{6-7}
\cmidrule{9-14}
\phantom{Var.}&\multicolumn{6}{c}{\phantom{Var.}}&&\multicolumn{2}{c}{\textbf{Task I}}&\multicolumn{2}{c}{\textbf{Task II}}&\multicolumn{2}{c}{\textbf{Task III}} \\
\cmidrule{9-14}
& {T $[s]\downarrow$} & {S $[\%]\uparrow$} & {T $[s]\downarrow$} & {S $[\%]\uparrow$} & {T $[s]\downarrow$} & {S $[\%]\uparrow$} & {} & {T $[s]\downarrow$} & {S $[\%]\uparrow$}& {T $[s]\downarrow$} & {S $[\%]\uparrow$}& {T $[s]\downarrow$} & {S $[\%]\uparrow$} \\
\toprule
\textbf{\#1}&$1.38{\pm}0.05$&$55$\%&$1.36{\pm}0.06$&$57$\%&$1.34{\pm}0.06$&$82$\%&&$1.38{\pm}0.05$&$35$\%&$1.39{\pm}0.02$&$32$\%&$1.41{\pm}0.05$&$36$\% \\
\midrule
\textbf{\#5}&$0.44{\pm}0.02$&$57$\%&$0.49{\pm}0.05$&$60$\%&$0.47{\pm}0.07$&$81$\%&&$0.47{\pm}0.03$&$30$\%&$0.46{\pm}0.06$&$34$\%&$0.47{\pm}0.05$&$37$\% \\
\midrule
\textbf{\#20}&$0.29{\pm}0.07$&$52$\%&$0.34{\pm}0.05$&$60$\%&$0.3{\pm}0.05$&$83$\%&&$0.3{\pm}0.04$&$35$\%&$0.28{\pm}0.01$&$32$\%&$0.28{\pm}0.01$&$\mathbf{40}$\% \\
\toprule
\textbf{Main (\#10)}&$\mathbf{0.22{\pm}0.01}$&$\mathbf{58}$\%&$\mathbf{0.23{\pm}0.05}$&$\mathbf{61.3}$\%&$\mathbf{0.24{\pm}0.04}$&$\mathbf{84.5}$\%&&$\mathbf{0.24{\pm}0.03}$&$\mathbf{35.7}$\%&$\mathbf{0.22{\pm} 0.01}$&$\mathbf{34.6}$\%&$\mathbf{0.22{\pm}0.01}$&$33.4$\% \\
\bottomrule
\end{tabular*}}
\end{center}
\end{table*}

\begin{figure}[htbp] 
\centering
\includegraphics[width=2.7 in]{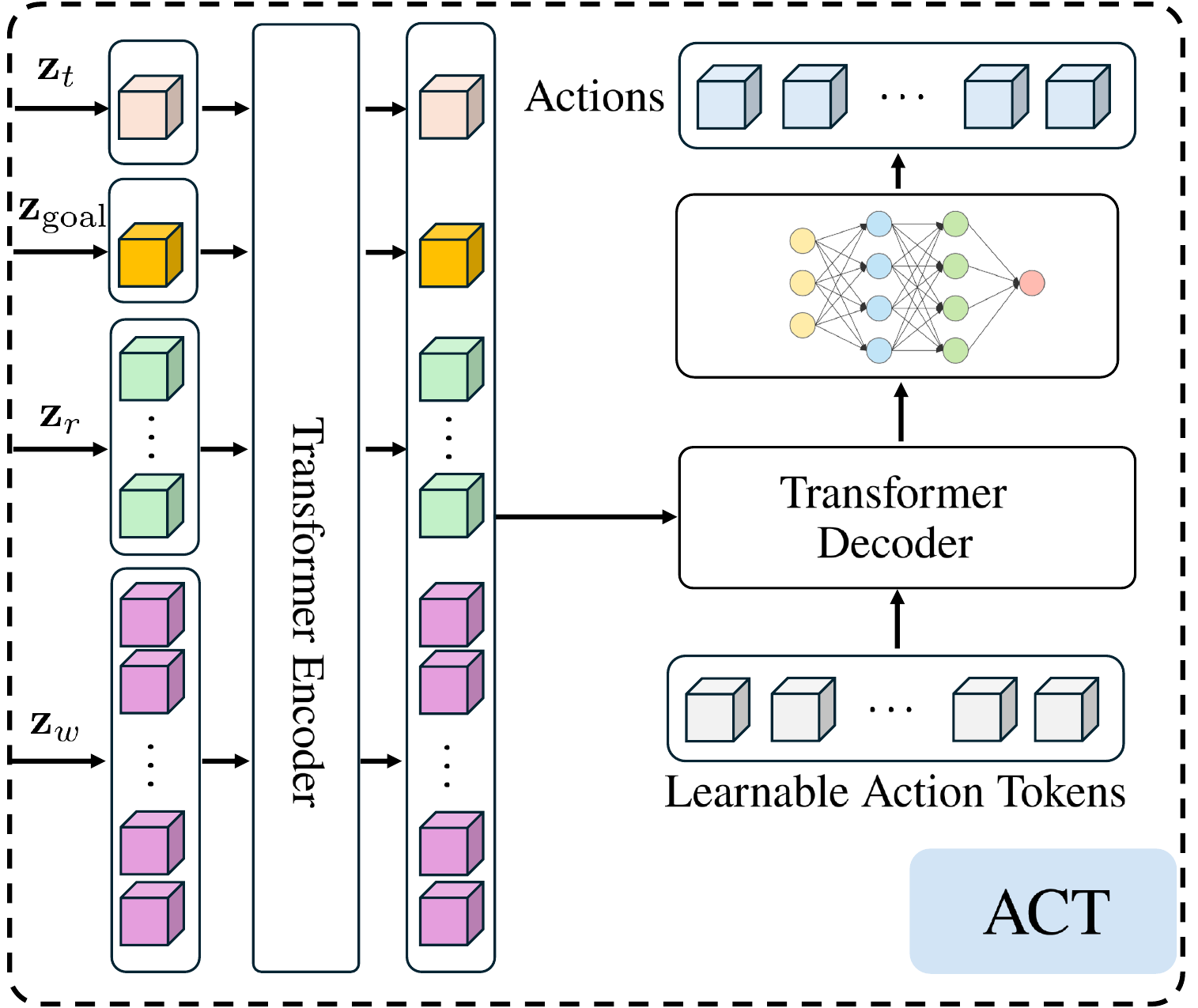}
\caption{\textbf{ACT Network Architecture} \cite{yang2025deep, zhao2023learning}. The framework leverages current-time-step workspace information (including the workspace and robot point clouds) together with configuration space features (current-time-step and goal configurations) through a standard pair-wise attention mechanism to generate an action sequence that guides the robotic manipulator toward the motion planning goal. $\mathbf{z}_t$, $\mathbf{z}_{\text{goal}}$, $\mathbf{z}_r$, and $\mathbf{z}_w$ are the current time-step configuration, planning goal configuration, robot point cloud, and scene point cloud embeddings, respectively.}
\label{fig_9 - act}
\end{figure}

\begin{figure}[htbp] 
\centering
\includegraphics[width=3 in]{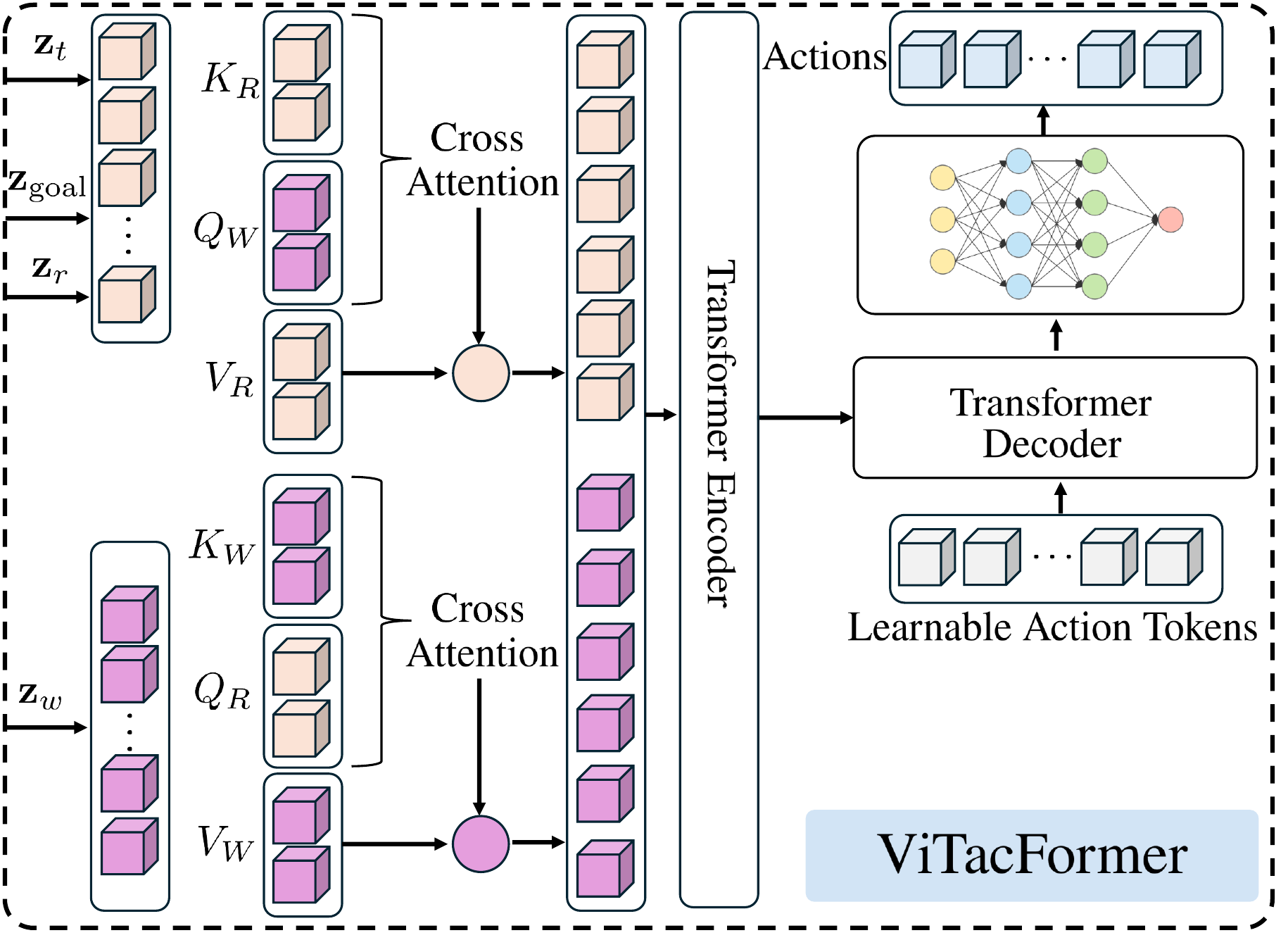}
\caption{\textbf{ViTacFormer Network Architecture}  \cite{heng2025vitacformer}. The framework utilizes a cross-attention mechanism to learn the interaction between planning sensing modalities. $\mathbf{z}_t$, $\mathbf{z}_{\text{goal}}$, $\mathbf{z}_r$, and $\mathbf{z}_w$ are the current time-step configuration, planning goal configuration, robot point cloud, and scene point cloud embeddings, respectively. $K_R$, $Q_R$, and $V_R$ are the robot embedding's key, query, and value matrices, respectively. $K_W$, $Q_W$, and $V_W$ are workspace embedding's key, query, and value matrices, respectively.}
\label{fig_10 - vitacformer}
\end{figure}

\vspace{0.2cm}
\noindent
\textbf{Combining Planning Modalities.} In this part, we compare the performance of M$\pi$NetsFusion against different frameworks with specific architectures for combining motion planning sensing modalities (i.e., robotic manipulator and workspace sensing information). These frameworks are as follows:

\vspace{0.2cm}
\noindent
\textit{Action Chunking Transformers (ACT)}:  This framework is inspired by \cite{zhao2023learning} and concurrent work \cite{yang2025deep}, which contains an encoder-decoder transformer architecture to predict feature joint angles given workspace and configuration space embeddings. This framework combines various planning sensing modalities through a vanilla attention mechanism for motion planning. We adopt this framework for the UR5e robotic manipulator to compare the performance with M$\pi$NetsFusion. Figure \ref{fig_9 - act} demonstrates the structure of this framework.

\begin{table}[htbp]
\centering
\caption{Average success rate and time comparison of M$\pi$NetsFusion with planners with different action chunk sizes across all held-out planning tasks.}
\label{tab: average_time_chunks}
\renewcommand{\arraystretch}{1.1}
%\resizebox{0.4\textwidth}{!}{%
\begin{tabular}{@{}p{0.15\linewidth}p{0.13\linewidth}p{0.07\linewidth}p{0.08\linewidth}p{0.15\linewidth}p{0.1\linewidth}@{}}
\toprule
&&\multicolumn{1}{c}{\textbf{\#1}}&\multicolumn{1}{c}{\textbf{\#5}}&\multicolumn{1}{c}{\textbf{\#20}}&\multicolumn{1}{c}{\textbf{Main (\#10)}} \\
\toprule
\multirow{6}{*}{$\frac{\text{T}}{\text{T}_{\text{M$\pi$NetsFusion}}}$}&\textbf{TableTop}&\multicolumn{1}{c}{$6.27$}&\multicolumn{1}{c}{$2.0$}&\multicolumn{1}{c}{$1.31$}&\multicolumn{1}{c}{$1.0$} \\ \hhline{~-----}
                                                              &\textbf{Box}&\multicolumn{1}{c}{$5.91$}&\multicolumn{1}{c}{$2.13$}&\multicolumn{1}{c}{$1.47$}&\multicolumn{1}{c}{$1.0$} \\  \hhline{~-----}
                                                              &\textbf{Bins}&\multicolumn{1}{c}{$5.58$}&\multicolumn{1}{c}{$1.95$}&\multicolumn{1}{c}{$1.25$}&\multicolumn{1}{c}{$1.0$} \\  \hhline{~-----}
                                                              &\textbf{Shelf I}&\multicolumn{1}{c}{$5.57$}&\multicolumn{1}{c}{$1.95$}&\multicolumn{1}{c}{$1.25$}&\multicolumn{1}{c}{$1.0$} \\  \hhline{~-----}
                                                              &\textbf{Shelf II}&\multicolumn{1}{c}{$6.31$}&\multicolumn{1}{c}{$2.09$}&\multicolumn{1}{c}{$1.27$}&\multicolumn{1}{c}{$1.0$} \\  \hhline{~-----}
                                                              &\textbf{Shelf III}&\multicolumn{1}{c}{$6.40$}&\multicolumn{1}{c}{$2.13$}&\multicolumn{1}{c}{$1.27$}&\multicolumn{1}{c}{$1.0$} \\ \hhline{~-----}
                                                              &\textbf{Average}&\multicolumn{1}{c}{$6$}&\multicolumn{1}{c}{$2.04$}&\multicolumn{1}{c}{$1.3$}&\multicolumn{1}{c}{$1.0$} \\
\bottomrule
\end{tabular}
\end{table}

\vspace{0.2cm}
\noindent
\textit{M$\pi$NetsFusion - Planning Embeddings Considered Separately}: In this case, we consider the current and goal configuration embeddings, the robot point cloud embedding, and workspace point cloud embedding as separate inputs to the transformer encoder such that these embeddings can only communicate through bottleneck tokens within the transformer encoder. As a result, the fusion transformer attends to four motion planning modalities instead of the original two.

\begin{figure}[htbp] 
\centering
\includegraphics[width=3 in]{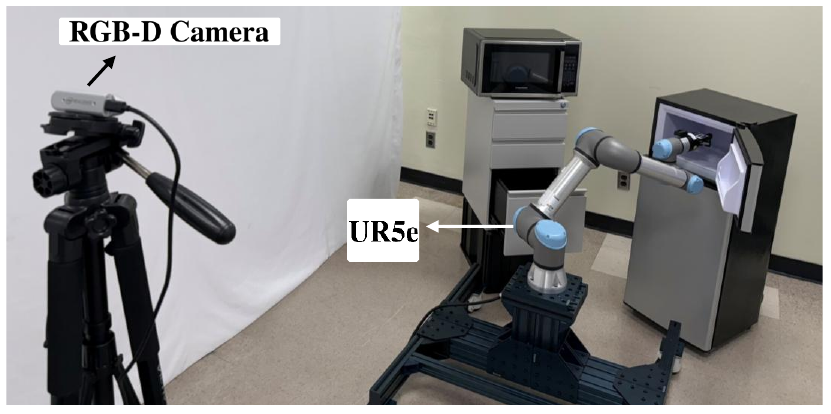}
\caption{\textbf{Experimental Setup} for real-world deployment of M$\pi$NetsFusion on UR5e robotic manipulator. The calibrated RGB-D camera is an Intel RealSense D435i used to capture workspace observations.}
\label{fig_11 - expsetup}
\end{figure}

\begin{figure}[htbp] 
\centering
\includegraphics[width=0.48\textwidth]{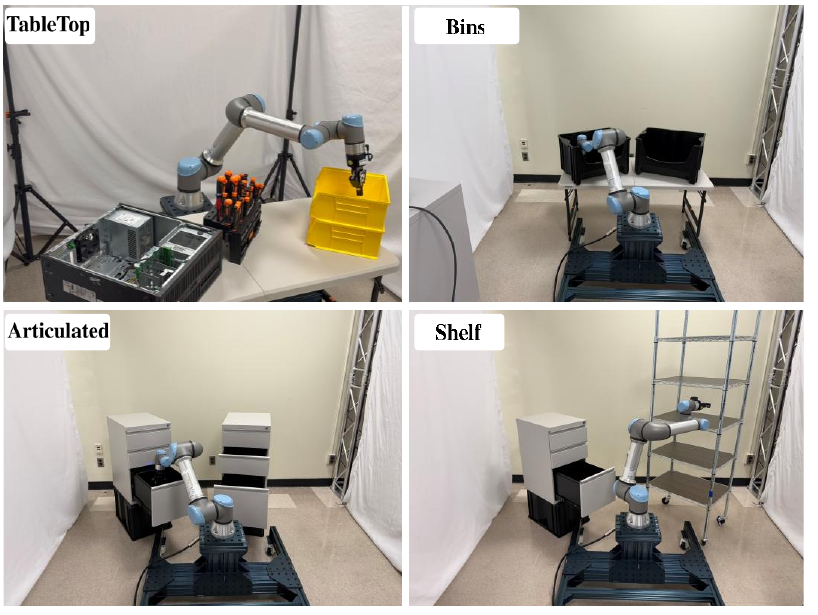}
\caption{\textbf{Evaluation Tasks} for M$\pi$NetsFusion real-world deployment.}
\label{fig_12 - exptasks}
\end{figure}

\vspace{0.2cm}
\noindent
\textit{M$\pi$NetsFusion - Configuration Space Embeddings Combined}: Instead of considering separate tokens for representing current and goal configuration embeddings, we first concatenate their embeddings and consider it as a single input to the transformer encoder. In this case, the fusion transformer attends to three motion planning modalities instead of the original two, which are configuration space embeddings, robot point cloud embedding, and workspace point cloud embedding, to communicate through bottleneck tokens.

\begin{figure*}[htbp] 
\centering
\includegraphics[width=0.8\textwidth]{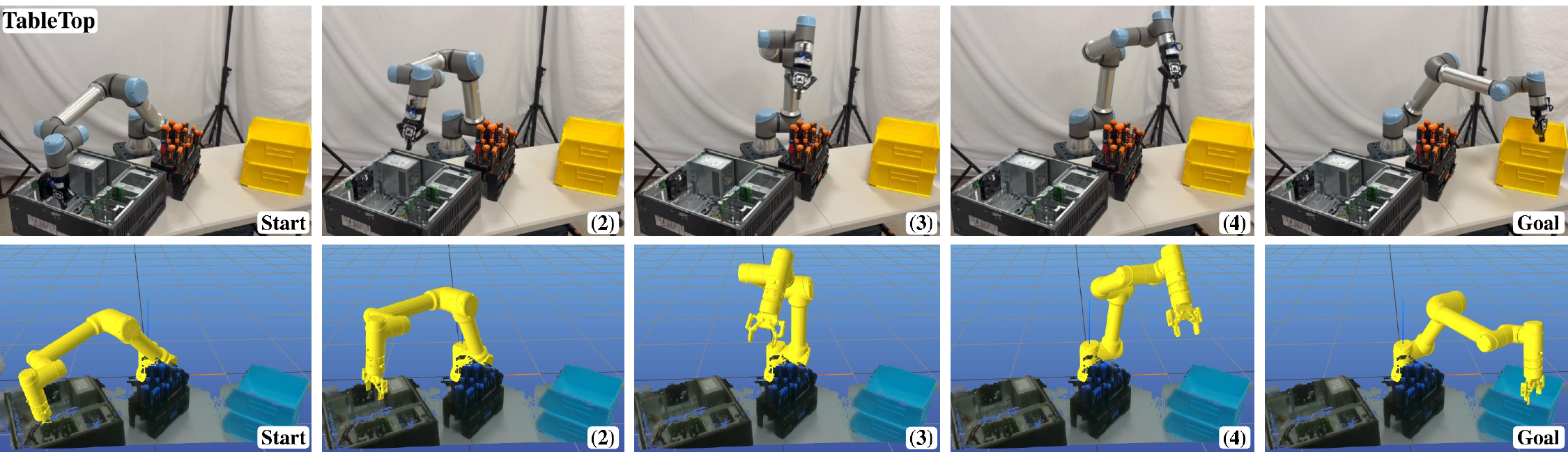}
\caption{\textbf{TableTop Task Evaluation} of M$\pi$NetsFusion in the real-world. The path profile from the start configuration to the goal configuration is demonstrated through frames 2-4 in the real-world (top), and simulated scene point cloud (bottom).}
\label{fig_13 - tabletopreal}
\end{figure*}

\begin{figure*}[t] 
\centering
\includegraphics[width=0.8\textwidth]{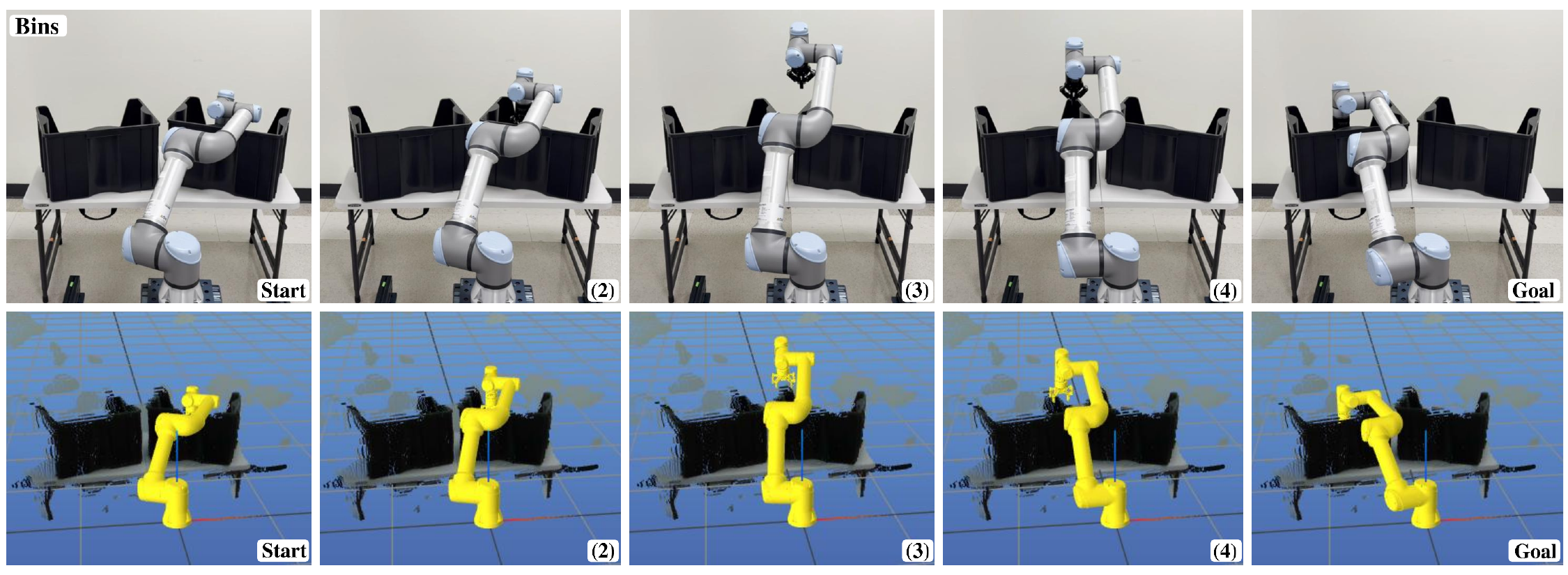}
\caption{\textbf{Bins Task Evaluation} of M$\pi$NetsFusion in the real-world. The path profile from the start configuration to the goal configuration is demonstrated through frames 2-4 in the real-world (top), and simulated scene point cloud (bottom).}
\label{fig_13- binreal}
\end{figure*}

\begin{figure*}[htbp] 
\centering
\includegraphics[width=0.8\textwidth]{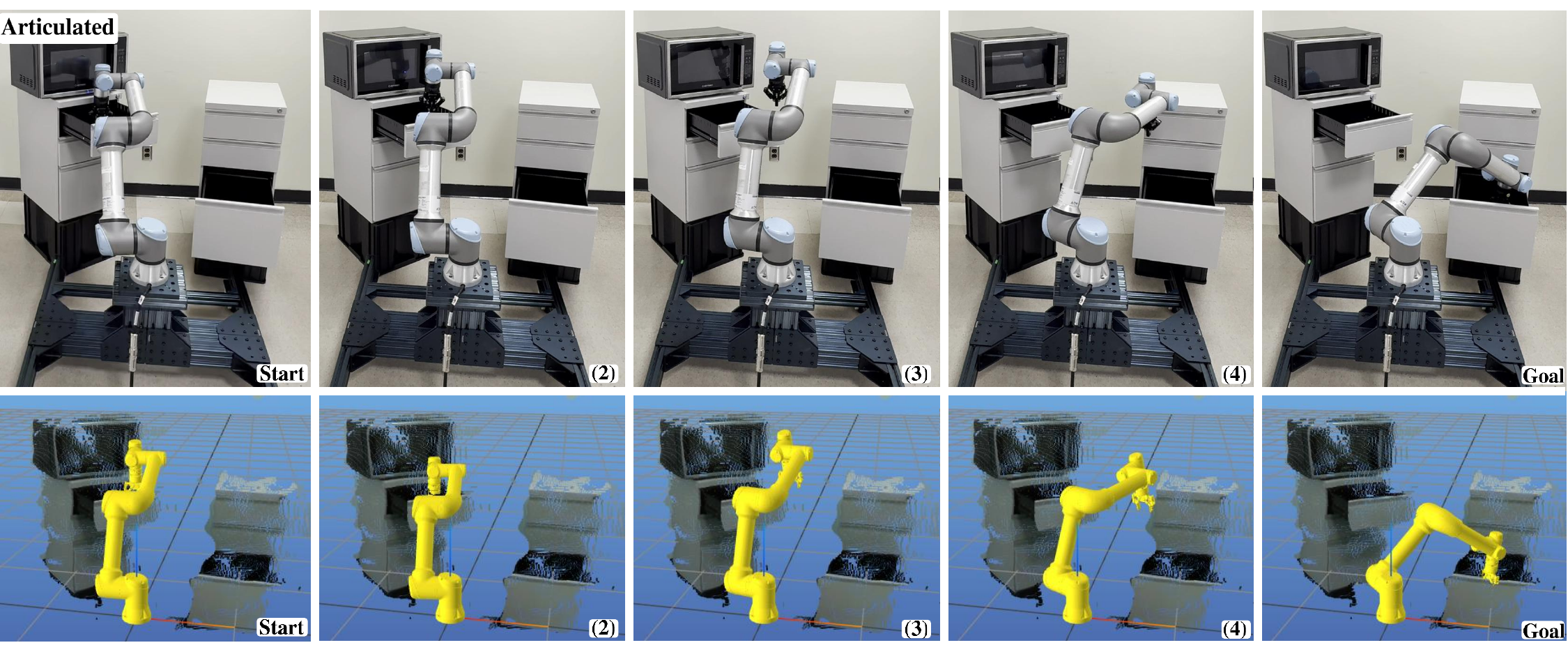}
\caption{\textbf{Articulated Task Evaluation} of M$\pi$NetsFusion in the real-world. The path profile from the start configuration to the goal configuration is demonstrated through frames 2-4 in the real-world (top), and simulated scene point cloud (bottom).}
\label{fig_14 - articulatedreal}
\end{figure*}

\vspace{0.2cm}
\noindent
\textit{ViTacFormer} \cite{heng2025vitacformer}: Originally proposed for learning cross-modal representations in visuo-tactile manipulation, this framework employs a cross-attention mechanism to model interactions between different sensing modalities. We adopt this approach to learn and encode the relationships among various planning sensing modalities. Similar to SIMPNet \cite{soleymanzadeh2025simpnet} and M$\pi$NetsFusion, we concatenate the current time step and goal configuration embeddings with robot point cloud embeddings as a single planning modality, and utilize the cross-attention mechanism to learn the interaction between these sensing modalities as shown in Figure \ref{fig_10 - vitacformer}.

For comparison, we report the success rate, reaching goal, and cold start time (CS) of each structure. Cold start time is defined as the average time to respond to a new problem, which is a metric to measure reactivity to a new planning problem.

M$\pi$NetsFusion outperforms ACT in both goal-reaching and success rate across all evaluation tasks, as shown in Table \ref{tab: ablation-frameworks}. ACT models the interaction between planning modalities using a standard attention mechanism, which often causes the workspace point-cloud embedding to dominate in evaluation static environments. In contrast, M$\pi$NetsFusion utilizes bottleneck tokens to learn interactions between robot and workspace planning embeddings, which leads to more efficient motion planning. M$\pi$NetsFusion also performs comparably to ACT in terms of cold-start time, as shown in Table \ref{tab: coldstart_time}, which is due to the fact that both models share the same network size.

$\pi$-Config and $\pi$-All are variants of M$\pi$NetsFusion that further separate the planning modalities, allowing them to communicate through bottleneck tokens rather than simple concatenation. As shown in Table \ref{tab: ablation-frameworks}, their performance is comparable to M$\pi$NetsFusion. However, both variants exhibit higher cold-start times, which can be attributed to their larger model sizes as demonstrated in Table \ref{tab: coldstart_time}.

The ViTacFormer framework also performs comparably to M$\pi$NetsFusion in terms of goal reaching, success rate, and cold-start time, as shown in Tables \ref{tab: ablation-frameworks} and \ref{tab: coldstart_time}. Its performance further supports the claim that explicitly processing robot-related and workspace-embedding signals enhances the efficiency of neural motion planners.

\begin{figure*}[htbp] 
\centering
\includegraphics[width=0.8\textwidth]{Figures/15ShelfReal.pdf}
\caption{\textbf{Shelf Task Evaluation} of M$\pi$NetsFusion in the real-world. The path profile from the start configuration to the goal configuration is demonstrated through frames 2-4 in the real-world (top), and simulated scene point cloud (bottom).}
\label{fig_15- shelfreal}
\end{figure*}

\vspace{0.2cm}
\noindent
\textbf{Action Chunk.} Since M$\pi$NetsFusion is an open-loop planner that unrolls the policy for a fixed number of steps, we vary the action-chunk size to assess its effect on planning accuracy. We evaluate chunk sizes of 1, 5, 10, and 20 to examine how different values influence performance. Because the planner operates in an open-loop, end-to-end manner, the rollout length is kept fixed across all experiments for consistency, such that models with action chunks of 1, 5, 10, and 20 are rolled out for 100, 20, 10, and 5 steps. Table \ref{tab: action_chunks} demonstrates the performance of the planner with different action chunk sizes across all evaluation planning tasks, while Table \ref{tab: average_time_chunks} provides average time comparison between these planners. 

As shown in Table \ref{tab: action_chunks}, M$\pi$NetsFusion with an action-chunk size of 10 achieves a success rate comparable to the policy version using a chunk size of 1. This is expected because the environments are static, and both planners produce trajectories of equal length under the max rollout assumption. However, the main framework is approximately six times faster than the model with an action chunk of size 1, as demonstrated in Table \ref{tab: average_time_chunks}. This improvement arises because the policy with a chunk size of 1 must sample the robot’s body point cloud in more steps, which results in higher planning times. 

A similar pattern is observed when comparing the policy with an action-chunk size of 5 to the main model. As shown in Table \ref{tab: action_chunks}, both achieve comparable success rates. However, the main policy is at least twice as fast as the policy with an action chunk size of 5, as demonstrated in Table \ref{tab: average_time_chunks}. This planning time difference arises since the policy with an action chunk of size 10 necessitates more point-cloud sampling on the body of the robotic manipulator during inference.

The main policy ($\pi$ \#10) also performs comparably to the model with an action chunk size of 20 ( see Table \ref{tab: action_chunks}), which is expected given the open-loop nature of the planners. However, the main policy is slightly faster (1.3×) than the policy with an action chunk size of 20, despite requiring fewer point cloud sampling steps on the body of the robotic manipulator. This performance gap can be traced back to the data-processing stage: when training with an action chunk size of 20, many dataset trajectory segments require post-pended zero-masked values for the delta-joint actions. These padded sequences increase the effective inference time of the motion planner.

\vspace{0.2cm}
\noindent
\textbf{Workspace Generation.} We evaluate the impact of different workspace generation strategies on the performance of the neural motion planner. Specifically, we train the planner on two datasets: one generated by \name and one generated by random workspace generation. Both datasets contain approximately 430K trajectories, and the network is trained for 150K gradient steps in each case. Table~\ref{tab: workspacegenerationablation} reports the planning performance across tasks for the two datasets.

\begin{table}[htbp]
\centering
\caption{Comparing \name and random workspace generation for motion planning.}
\label{tab: workspacegenerationablation}
\resizebox{0.49\textwidth}{!}{%
\begin{tabular}{@{}p{0.20\linewidth}p{0.10\linewidth}p{0.1\linewidth}p{0.1\linewidth}p{0.1\linewidth}p{0.1\linewidth}p{0.1\linewidth}@{}}
\toprule
\multicolumn{1}{l}{\phantom{Var.}}&\multicolumn{1}{c}{\scriptsize{\textbf{TableTop}}}&\multicolumn{1}{c}{\scriptsize{\textbf{Box}}}&\multicolumn{1}{c}{\scriptsize{\textbf{Bins}}}&\multicolumn{1}{c}{\scriptsize{\textbf{Shelf1}}}&\multicolumn{1}{c}{\scriptsize{\textbf{Shelf2}}}&\multicolumn{1}{c}{\scriptsize{\textbf{Shelf3}}} \\
\toprule
\multicolumn{1}{l}{\scriptsize{\textbf{M$\pi$NetsFusion-Random}}}&\multicolumn{1}{c}{\scriptsize{45\%}}&\multicolumn{1}{c}{\scriptsize{55\%}}&\multicolumn{1}{c}{\scriptsize{71\%}}&\multicolumn{1}{c}{\scriptsize{33\%}}&\multicolumn{1}{c}{\scriptsize{26\%}}&\multicolumn{1}{c}{\scriptsize{32\%}} \\
\toprule
\multicolumn{1}{l}{\scriptsize{\textbf{M$\pi$NetsFusion-\name}}}&\multicolumn{1}{c}{\scriptsize{\textbf{48\%}}}&\multicolumn{1}{c}{\scriptsize{\textbf{59\%}}}&\multicolumn{1}{c}{\scriptsize{\textbf{77\%}}}&\multicolumn{1}{c}{\scriptsize{\textbf{34\%}}}&\multicolumn{1}{c}{\scriptsize{\textbf{29\%}}}&\multicolumn{1}{c}{\scriptsize{\textbf{37\%}}} \\
\bottomrule
\end{tabular}}
\end{table}

The results show that \name improves success rates across all planning tasks. However, performance on shelf tasks remains around 30\% for both workspace generation approaches and dataset sizes, likely due to the limited diversity of shelf primitives in the generation pipeline.

\vspace{0.2cm}
\noindent
\textbf{Bottleneck Tokens.} We also evaluate the performance of M$\pi$NetsFusion with different numbers of bottleneck tokens. All models are trained with the original M$\pi$NetsFusion dataset. The results are listed in Table~\ref{tab: numbottlenecks}.

\begin{table}[htbp]
\centering
\caption{Planning comparison between M$\pi$NetsFusion and ablation frameworks with different numbers of bottleneck tokens in terms of success rate.}
\label{tab: numbottlenecks}
\resizebox{0.4\textwidth}{!}{%
\begin{tabular}{@{}p{0.20\linewidth}p{0.10\linewidth}p{0.1\linewidth}p{0.1\linewidth}p{0.1\linewidth}p{0.1\linewidth}p{0.1\linewidth}@{}}
\toprule
\multicolumn{1}{l}{\phantom{Var.}}&\multicolumn{1}{c}{\scriptsize{\textbf{TableTop}}}&\multicolumn{1}{c}{\scriptsize{\textbf{Box}}}&\multicolumn{1}{c}{\scriptsize{\textbf{Bins}}}&\multicolumn{1}{c}{\scriptsize{\textbf{Shelf1}}}&\multicolumn{1}{c}{\scriptsize{\textbf{Shelf2}}}&\multicolumn{1}{c}{\scriptsize{\textbf{Shelf3}}} \\
\toprule
\multicolumn{1}{l}{\scriptsize{\textbf{ACT (w/o)}}}&\multicolumn{1}{c}{\scriptsize{37\%}}&\multicolumn{1}{c}{\scriptsize{51\%}}&\multicolumn{1}{c}{\scriptsize{65.25\%}}&\multicolumn{1}{c}{\scriptsize{30.6\%}}&\multicolumn{1}{c}{\scriptsize{27\%}}&\multicolumn{1}{c}{\scriptsize{28.3\%}} \\
\toprule
\multicolumn{1}{l}{\scriptsize{\textbf{\#1}}}&\multicolumn{1}{c}{\scriptsize{54\%}}&\multicolumn{1}{c}{\scriptsize{60\%}}&\multicolumn{1}{c}{\scriptsize{84\%}}&\multicolumn{1}{c}{\scriptsize{34\%}}&\multicolumn{1}{c}{\scriptsize{32\%}}&\multicolumn{1}{c}{\scriptsize{35\%}} \\
\toprule
\multicolumn{1}{l}{\scriptsize{\textbf{\#2}}}&\multicolumn{1}{c}{\scriptsize{54\%}}&\multicolumn{1}{c}{\scriptsize{61\%}}&\multicolumn{1}{c}{\scriptsize{80\%}}&\multicolumn{1}{c}{\scriptsize{\textbf{38\%}}}&\multicolumn{1}{c}{\scriptsize{33\%}}&\multicolumn{1}{c}{\scriptsize{32\%}} \\
\toprule
\multicolumn{1}{l}{\scriptsize{\textbf{Main (\#4)}}}&\multicolumn{1}{c}{\scriptsize{\textbf{58\%}}}&\multicolumn{1}{c}{\scriptsize{\textbf{61.3\%}}}&\multicolumn{1}{c}{\scriptsize{84.5\%}}&\multicolumn{1}{c}{\scriptsize{35.7\%}}&\multicolumn{1}{c}{\scriptsize{\textbf{34.6\%}}}&\multicolumn{1}{c}{\scriptsize{33.4\%}} \\
\toprule
\multicolumn{1}{l}{\scriptsize{\textbf{\#6}}}&\multicolumn{1}{c}{\scriptsize{56\%}}&\multicolumn{1}{c}{\scriptsize{61\%}}&\multicolumn{1}{c}{\scriptsize{\textbf{86\%}}}&\multicolumn{1}{c}{\scriptsize{32\%}}&\multicolumn{1}{c}{\scriptsize{32\%}}&\multicolumn{1}{c}{\scriptsize{\textbf{38\%}}} \\
\toprule
\multicolumn{1}{l}{\scriptsize{\textbf{\#8}}}&\multicolumn{1}{c}{\scriptsize{53\%}}&\multicolumn{1}{c}{\scriptsize{61\%}}&\multicolumn{1}{c}{\scriptsize{84\%}}&\multicolumn{1}{c}{\scriptsize{33\%}}&\multicolumn{1}{c}{\scriptsize{32\%}}&\multicolumn{1}{c}{\scriptsize{35\%}} \\
\bottomrule
\end{tabular}}
\end{table}

The results show that constraining information flow through bottleneck tokens prevents the scene point cloud from dominating the planning process, as demonstrated in the ablation study on combining planning modalities in Table~\ref{tab: ablation-frameworks}.

\vspace{0.2cm}
\noindent
\textbf{SetAbstration Radius:} We evaluate the performance of M$\pi$NetsFusion under different set abstraction radii in PointNet++~\cite{qi2017pointnet++}. The results are reported in Table~\ref{tab: saradius}. For this ablation, all models are trained on a dataset of approximately 430K trajectories for 150K gradient steps.

\begin{table}[htbp]
\centering
\caption{M$\pi$NetsFusion performance across various SetAbstraction radii.}
\label{tab: saradius}
\resizebox{0.45\textwidth}{!}{%
\begin{tabular}{@{}p{0.20\linewidth}p{0.10\linewidth}p{0.1\linewidth}p{0.1\linewidth}p{0.1\linewidth}p{0.1\linewidth}p{0.1\linewidth}@{}}
\toprule
\multicolumn{1}{l}{\phantom{Var.}}&\multicolumn{1}{c}{\scriptsize{\textbf{TableTop}}}&\multicolumn{1}{c}{\scriptsize{\textbf{Box}}}&\multicolumn{1}{c}{\scriptsize{\textbf{Bins}}}&\multicolumn{1}{c}{\scriptsize{\textbf{Shelf1}}}&\multicolumn{1}{c}{\scriptsize{\textbf{Shelf2}}}&\multicolumn{1}{c}{\scriptsize{\textbf{Shelf3}}} \\
\toprule
\multicolumn{1}{l}{\scriptsize{\textbf{Radius=$0.05$}}}&\multicolumn{1}{c}{\scriptsize{46\%}}&\multicolumn{1}{c}{\scriptsize{59.7\%}}&\multicolumn{1}{c}{\scriptsize{\textbf{77.6}\%}}&\multicolumn{1}{c}{\scriptsize{31.7\%}}&\multicolumn{1}{c}{\scriptsize{\textbf{29.7\%}}}&\multicolumn{1}{c}{\scriptsize{35.7\%}} \\
\toprule
\multicolumn{1}{l}{\scriptsize{\textbf{Radius=$0.1$ (Main)}}}&\multicolumn{1}{c}{\scriptsize{\textbf{48\%}}}&\multicolumn{1}{c}{\scriptsize{59\%}}&\multicolumn{1}{c}{\scriptsize{77\%}}&\multicolumn{1}{c}{\scriptsize{34\%}}&\multicolumn{1}{c}{\scriptsize{29\%}}&\multicolumn{1}{c}{\scriptsize{\textbf{37\%}}} \\
\toprule
\multicolumn{1}{l}{\scriptsize{\textbf{Radius=$0.3$}}}&\multicolumn{1}{c}{\scriptsize{\textbf{48\%}}}&\multicolumn{1}{c}{\scriptsize{\textbf{61\%}}}&\multicolumn{1}{c}{\scriptsize{74.4\%}}&\multicolumn{1}{c}{\scriptsize{\textbf{37.3\%}}}&\multicolumn{1}{c}{\scriptsize{28.7\%}}&\multicolumn{1}{c}{\scriptsize{36.7\%}} \\
\bottomrule
\end{tabular}}
\end{table}

The results show that planner performance varies only slightly across different set abstraction radii. This is because the downsampled point clouds (16 for the robot and 128 for the workspace) provide sufficient coverage of the planning space across the tested radius range.

\vspace{8pt}
\section{Real-world Deployment} \label{sec: experiments}
In this section, we evaluate our motion policy on a UR5e robotic manipulator in real-world environments. One or two calibrated Intel RealSense D435i RGB-D cameras are used to capture point cloud data of the environment, as illustrated in Figure~\ref{fig_11 - expsetup}. 

The real-world tasks follow the structure of some held-out evaluation tasks described in Section \ref{sec:main_eval} as demonstrated in Figure \ref{fig_12 - exptasks}. We categorize these tasks into four groups: the \textbf{TableTop Task}, the \textbf{Bin Task}, \textbf{Articulated Task}, and \textbf{Shelf Task}. Table \ref{tab: experiments} summarizes the planning results for these real-world tasks, and Figures \ref{fig_13 - tabletopreal}, \ref{fig_13- binreal}, \ref{fig_14 - articulatedreal}, and \ref{fig_15- shelfreal} demonstrate planning examples for each task category.

\begin{table}[htbp]
\centering
\caption{M$\pi$NetsFusion performance in real-world evaluation tasks, reported using \textbf{ReachedGoal} and \textbf{Successful} metrics.}
\label{tab: experiments}
\renewcommand{\arraystretch}{1.1}
\resizebox{0.45\textwidth}{!}{%
\begin{tabular}{@{}p{0.20\linewidth}p{0.10\linewidth}p{0.20\linewidth}p{0.10\linewidth}p{0.10\linewidth}p{0.20\linewidth}p{0.20\linewidth}@{}}
\toprule
\rowcolor{gray!20}\multicolumn{1}{l}{\phantom{Var.}}&\multicolumn{6}{c}{\textbf{TableTop}}\\
\midrule
\multicolumn{1}{l}{\phantom{Var.}}&\multicolumn{1}{c}{\textbf{Scenario 1}}&\multicolumn{1}{c}{\textbf{Scenario 2}}&\multicolumn{1}{c}{\textbf{Scenario 3}}&\multicolumn{1}{c}{\textbf{Scenario 4}}&\multicolumn{1}{c}{\textbf{Scenario 5}}&\multicolumn{1}{|c}{\textbf{Total \#}} \\
\midrule
\multicolumn{1}{l}{\textbf{ReachedGoal} $\uparrow$}&\multicolumn{1}{c}{10/10}&\multicolumn{1}{c}{10/10}&\multicolumn{1}{c}{10/10}&\multicolumn{1}{c}{10/10}&\multicolumn{1}{c}{10/10}&\multicolumn{1}{|c}{50/50} \\
\midrule
\multicolumn{1}{l}{\textbf{Successful} $\uparrow$}&\multicolumn{1}{c}{10/10}&\multicolumn{1}{c}{9/10}&\multicolumn{1}{c}{10/10}&\multicolumn{1}{c}{9/10}&\multicolumn{1}{c}{10/10}&\multicolumn{1}{|c}{48/50} \\
\midrule
\rowcolor{gray!20}\multicolumn{1}{l}{\phantom{Var.}}&\multicolumn{6}{c}{\textbf{Bins}}\\
\midrule
\multicolumn{1}{l}{\phantom{Var.}}&\multicolumn{1}{c}{\textbf{Scenario 1}}&\multicolumn{1}{c}{\textbf{Scenario 2}}&\multicolumn{1}{c}{\textbf{Scenario 3}}&\multicolumn{1}{c}{\textbf{Scenario 4}}&\multicolumn{1}{c}{\textbf{Scenario 5}}&\multicolumn{1}{|c}{\textbf{Total \#}} \\
\midrule
\multicolumn{1}{l}{\textbf{ReachedGoal} $\uparrow$}&\multicolumn{1}{c}{10/10}&\multicolumn{1}{c}{10/10}&\multicolumn{1}{c}{10/10}&\multicolumn{1}{c}{10/10}&\multicolumn{1}{c}{10/10}&\multicolumn{1}{|c}{50/50} \\
\midrule
\multicolumn{1}{l}{\textbf{Successful} $\uparrow$}&\multicolumn{1}{c}{7/10}&\multicolumn{1}{c}{8/10}&\multicolumn{1}{c}{9/10}&\multicolumn{1}{c}{10/10}&\multicolumn{1}{c}{8/10}&\multicolumn{1}{|c}{42/50} \\
\midrule
\rowcolor{gray!20}\multicolumn{1}{l}{\phantom{Var.}}&\multicolumn{6}{c}{\textbf{Articulated}}\\
\midrule
\multicolumn{1}{l}{\phantom{Var.}}&\multicolumn{1}{c}{\textbf{Scenario 1}}&\multicolumn{1}{c}{\textbf{Scenario 2}}&\multicolumn{1}{c}{\textbf{Scenario 3}}&\multicolumn{1}{c}{\textbf{Scenario 4}}&\multicolumn{1}{c}{\textbf{Scenario 5}}&\multicolumn{1}{|c}{\textbf{Total \#}} \\
\midrule
\multicolumn{1}{l}{\textbf{ReachedGoal} $\uparrow$}&\multicolumn{1}{c}{10/10}&\multicolumn{1}{c}{10/10}&\multicolumn{1}{c}{10/10}&\multicolumn{1}{c}{10/10}&\multicolumn{1}{c}{10/10}&\multicolumn{1}{|c}{50/50} \\
\midrule
\multicolumn{1}{l}{\textbf{Successful} $\uparrow$}&\multicolumn{1}{c}{7/10}&\multicolumn{1}{c}{7/10}&\multicolumn{1}{c}{5/10}&\multicolumn{1}{c}{6/10}&\multicolumn{1}{c}{5/10}&\multicolumn{1}{|c}{30/50} \\
\midrule
\rowcolor{gray!20}\multicolumn{1}{l}{\phantom{Var.}}&\multicolumn{6}{c}{\textbf{Shelf}}\\
\midrule
\multicolumn{1}{l}{\phantom{Var.}}&\multicolumn{1}{c}{\textbf{Scenario 1}}&\multicolumn{1}{c}{\textbf{Scenario 2}}&\multicolumn{1}{c}{\textbf{Scenario 3}}&\multicolumn{1}{c}{\textbf{Scenario 4}}&\multicolumn{1}{c}{\textbf{Scenario 5}}&\multicolumn{1}{|c}{\textbf{Total \#}} \\
\midrule
\multicolumn{1}{l}{\textbf{ReachedGoal} $\uparrow$}&\multicolumn{1}{c}{10/10}&\multicolumn{1}{c}{10/10}&\multicolumn{1}{c}{10/10}&\multicolumn{1}{c}{10/10}&\multicolumn{1}{c}{10/10}&\multicolumn{1}{|c}{50/50} \\
\midrule
\multicolumn{1}{l}{\textbf{Successful} $\uparrow$}&\multicolumn{1}{c}{3/10}&\multicolumn{1}{c}{6/10}&\multicolumn{1}{c}{4/10}&\multicolumn{1}{c}{2/10}&\multicolumn{1}{c}{3/10}&\multicolumn{1}{|c}{18/50} \\
\bottomrule
\end{tabular}}
\end{table}

For the table-top tasks, the planner reaches all the planning goals (50/50), among which 48 (96\%) are successful. For the bins tasks, the planner reaches the planning goal in all the planning problems (50/50), among which 42 (84\%) are successful and collision-free. These results are consistent with the simulation performance on the bins tasks. For the articulated tasks, the planner also reaches the planning goal in all the planning problems (50/50), among which 30 (60\%) are successful and collision-free. For the shelf task, the planner reaches the planning goal in all the planning problems (50/50); however, only 18 out of 50 (36\%) are successful and collision-free (30\%). These results are also consistent with the simulation performance on the shelf tasks. This is primarily due to the limited diversity of shelf-like primitives in the training dataset. Additionally, in real-world deployment, workspace reconstruction relies on point clouds from up to two RGB-D cameras, which results in partial and noisy observations. In contrast, the network is trained on complete workspace point clouds sampled directly from primitive surfaces. This sim-to-real perception gap degrades performance in narrow spaces in shelf environments.
\section{Conclusions} \label{sec: conclusion}
In this paper, we present \textbf{PerFACT}, which contains two modules to learn the motion policy for robotic manipulators. The first module, \textbf{\name}, is an LLM-based workspace generation framework that is used for collecting large-scale motion planning datasets. The second module, \textbf{M$\pi$NetsFusion}, is a novel neural motion planning framework for end-to-end motion planning for robotic manipulators. MotionGeneralizer has specific features that make it capable of generating motion-planning datasets for any robotic manipulator with arbitrary degrees of freedom (DOF). These features are as follows:

\vspace{0.2cm}
\noindent
\textbf{Diverse Workspace generation}: \name leverages procedural everyday geometric primitives generation with the semantic reasoning capabilities of LLMs to create feasible and cluttered workspaces around robotic manipulators for motion planning. This framework utilizes a chain-of-thought (CoT) prompting technique to create workspaces with high diversity.

\vspace{0.2cm}
\noindent
\textbf{Robot-agnostic, scene-specific motion planning problems}: \name samples collision-free, robot-agnostic poses within each workspace by considering the positions and orientations of all primitives. It then uses robot-specific inverse kinematics to compute feasible start and goal configurations for the selected robotic manipulator. The resulting planning problems can be solved by any off-the-shelf classical motion planner (e.g., OMPL \cite{sucan2012open} or cuRobo \cite{sundaralingam2023curobo}) for large-scale planning dataset collection.

\vspace{0.2cm}
\noindent
\textbf{Sensing modality}: \name samples points directly on the meshes of both the workspace primitives and the robotic manipulator at arbitrary configurations, to provide planners privileged access to accurate geometric and shape information. Compared to using RGB-D cameras in simulation, this approach is faster and provides a complete representation of the planning workspace.

We then utilized the \name framework to collect a large-scale dataset for training M$\pi$NetsFusion, a novel end-to-end, open-loop motion planner for the UR5e robotic manipulators. M$\pi$NetsFusion employs a bottleneck-based fusion transformer architecture that intelligently attends to various planning modalities. The model adopts an action-chunking transformer encoder–decoder architecture to predict future delta joint angles from the current time-step workspace and robot embedding information. The encoder is designed so that the robot and workspace embeddings communicate only through bottleneck tokens, which aggregate and balance the planning information while preventing the workspace embedding from dominating the planning signal.

We compare the performance of the proposed planner with state-of-the-art sampling-based and neural motion planners. The evaluation results demonstrate that M$\pi$NetsFusion achieves superior average planning time and comparable success rate relative to benchmark methods across all held-out planning tasks. Our ablation studies also demonstrate that treating the robotic manipulator and scene embeddings separately-instead of simple concatenation-yields substantial performance improvements. Additional separation of the robotic manipulator's configuration space and workspace information yields only marginal performance gains while increasing model size and cold-start time. Additionally, we showcase that action chunking significantly reduces the overall planning time due to the end-to-end nature of the proposed framework. Finally, our experimental results highlight the planner’s performance across a range of real-world planning tasks.

However, our proposed \name and M$\pi$NetsFusion have several limitations that will be potential research directions for our future work. \name relies on prompting an LLM to describe each workspace for similarity estimation. However, similarity between language descriptions of generated workspaces does not necessarily reflect the true geometric similarity between workspaces. Vision-based similarity metrics have the potential to provide more accurate similarity detection measures and could further improve \name to generate more diverse workspaces. Also, this framework only considers articulated primitives for workspace creation. Incorporating existing 3D datasets like Objaverse \cite{deitke2023objaverse} to sample simple everyday meshes to place between, within, or on top of the generated articulated primitives would produce more realistic and cluttered environments. Additionally, incorporating a broader set of articulated primitives would improve workspace realism and enhance the performance of downstream neural planners.

Also, M$\pi$NetsFusion is currently limited to static environments due to its end-to-end, open-loop design. To extend its applicability to dynamic environments, a high-frequency reactive motion planner like Riemannian Motion Policies (RMPs) \cite{ratliff2018riemannian} or Geometric Fabrics \cite{van2022geometric} would need to operate on top of M$\pi$NetsFusion to handle environmental changes in real time. Additionally, although \name enables the collection of motion planning data for robotic manipulators with arbitrary degrees of freedom, M$\pi$NetsFusion is trained and evaluated only for the UR5e manipulator. Thus, the proposed motion planner is embodiment-specific and cannot be directly applied to other robotic manipulators with different degrees of freedom. Moreover, the open-loop, end-to-end nature of M$\pi$NetsFusion can lead to minor collisions in narrow and cluttered environments. Following concurrent work~\cite{yang2025deep}, a reactive local motion planner such as Geometric Fabrics~\cite{van2022geometric} or Deep Refinement Learning (DRL) frameworks~\cite{li2021general, liu2024robotic}, can be used to refine the predicted trajectories and mitigate such collisions.
\bibliographystyle{IEEEtran}
\bibliography{ref}

\end{document}